\documentclass[11pt]{article}

\usepackage[a4paper,margin=1in]{geometry}
\usepackage{amsmath,amssymb}
\usepackage{graphicx}
\usepackage{float}
\usepackage{booktabs}
\usepackage{longtable}
\usepackage{tabularx}
\usepackage{array}
\usepackage[svgnames]{xcolor}
\usepackage{tikz}

\usetikzlibrary{arrows.meta,positioning}
\usepackage[hyphens]{url}
\usepackage[hidelinks]{hyperref}

\newcommand{\reportfig}[2][]{\includegraphics[#1]{#2}}

\title{{ WAM is capable in BEHAVIOR-1K: the 2nd-place solution to the 2026 BEHAVIOR Challenge}}

\author{%
Yelin He\textsuperscript{1}, Jiaquan Ye\textsuperscript{1}, Xinming Wei\textsuperscript{1,*}, Wenjie Xiao\textsuperscript{2}, \\ Gengming Zhang\textsuperscript{1}, Guoting Wei\textsuperscript{1},  Xianbiao Qi\textsuperscript{1, \dag}, Rong Xiao\textsuperscript{1}%
}

\date{}

\begin{document}
\maketitle
\begingroup
\renewcommand{\thefootnote}{}
\footnotetext{\textsuperscript{1}Shenzhen Intellifusion Technologies Co., Ltd., China. \textsuperscript{2}Johns Hopkins University. \dag: Xianbiao Qi contributed to the project when he worked at Intellifusion.
* denotes the corresponding author: Xinming Wei \textless\href{mailto:xinming.wei20@gmail.com}{xinming.wei20@gmail.com}\textgreater.}
\endgroup

\begin{abstract}
{
This work presents the 2nd-place solution (Team IF1001) to the 2026 BEHAVIOR Challenge, which evaluates mobile manipulation across 100 household tasks. Our method is built on a powerful world action model (WAM), Cosmos 3. 
We improve Cosmos 3 by using richer visual history, task-progress conditioning, and task-group and skill-specific LoRA adapters. During inference, we apply a separate progress estimator to supply task-progress context, and use task- and progress-conditioned routing to select a task-group or skill-specific policy, complemented by action-candidate aggregation, rolling inpainting, and task-specific temporal and gripper control. To meet the minimum inference requirement of 1~FPS, we use neither recursive self-improvement (RSI) nor agentic harness methods and rely on no closed-source models; during official private-test evaluation, our system handled concurrent requests across 100 evaluation ports with limited hardware resources, sustaining an average end-to-end inference rate above 2~FPS per port. Our method achieves a mean Q-score of 0.613 and a full-task success rate of 40.9\%, averaged over 3 self-evaluation rounds across all 100 tasks. On the official private test set, with 20 instances per task, our system achieves a mean Q-score of 0.615 and a full-task success rate of 41.2\%. 
}

\end{abstract}

\section{Introduction}

{ Building a robot with long-horizon household capabilities is a core goal in embodied AI. A household task generally requires navigation between rooms, coordinated manipulation of several objects, and a sequence of operations whose success depends on earlier actions. Small control errors can accumulate, and similar observations may call for different actions at different stages. The BEHAVIOR Challenge provides a setting for studying these difficulties through long-horizon activities in realistic household environments \cite{li2024,behavior2026}.

The leading solutions to the 2025 BEHAVIOR Challenge build on the $\pi_{0.5}$ vision-language-action (VLA) model and demonstrate the importance of policy adaptation through training data, task-progress estimation, and execution strategies \cite{bai2025,larchenko2025}. However, due to limited training data and generalization capability, VLA models often suffer from poor instruction-following and spatial-understanding abilities.

In this work, we present a WAM-based method, the 2nd-place solution, to the 2026 BEHAVIOR Challenge. In this solution, we

\begin{itemize}
    \item adopt Cosmos 3~\cite{nvidia2026} as our WAM backbone. We choose Cosmos 3 as our foundation model because it jointly models understanding and generation and has undergone large-scale pretraining. Cosmos 3 possesses strong language and visual reasoning capabilities while being capable of generating high-fidelity videos and actions.
    \item enhance Cosmos 3 by integrating richer visual history, task-progress estimation, and task-group LoRA fine-tuning to address observational ambiguity at complementary levels. History and progress estimation help distinguish visually similar observations possibly encountered at different task stages, while task grouping separates tasks with similar initial scene configurations but different required continuations. To improve manipulation reliability, we train separate LoRA adapters for picking, placing, opening, and closing on skill-specific demonstration segments pooled across tasks. Skill predictions from the progress estimator guide the router in selecting these skill-specific LoRA adapters.
    \item improve the inference pipeline by combining task-progress estimation, action-candidate aggregation, and rolling inpainting with task-specific execution windows, temporal retiming, and gripper-aware control.
    \item evaluate the main design choices and system configurations. Specifically, we evaluate the base policy, task-group LoRA fine-tuning, and several execution configurations, and report both partial goal completion and full-task success. The comparisons characterize the performance of the assembled system; they do not isolate the contribution of every component. 
\end{itemize}

\textit{In conclusion, this work demonstrates that WAM is a capable solution and has considerable potential for long-horizon household tasks such as those in the BEHAVIOR Challenge.}
}

\section{Related Work}

\subsection{BEHAVIOR and related benchmarks}
A variety of benchmarks~\cite{mees2022calvin, liu2023libero, chen2025robotwin, chen2026robodojo, huang2026robocoliseum, robochallenge2025, atreya2025roboarena, shang2026worldarena} have been developed to assess different aspects of robotic manipulation and embodied intelligence.
CALVIN~\cite{mees2022calvin} evaluates long-horizon language-conditioned manipulation, while LIBERO~\cite{liu2023libero} studies lifelong knowledge transfer across spatial, object, and goal variations. RoboTwin 2.0~\cite{chen2025robotwin} tests bimanual manipulation robustness under domain randomization. RoboDojo~\cite{chen2026robodojo} pairs real-robot evaluation with simulation tests of generalization, memory, precision, long-horizon execution, and open-vocabulary instruction following. RoboColiseum~\cite{huang2026robocoliseum} assesses instruction following, spatial reasoning, general manipulation, and robustness. RoboChallenge~\cite{robochallenge2025} hosts real-robot evaluation through Table30, whereas RoboArena~\cite{atreya2025roboarena} ranks real-robot policies using distributed, double-blind pairwise preferences. WorldArena~\cite{shang2026worldarena} evaluates embodied world models' perceptual quality and utility for data generation, policy evaluation, and action planning.

BEHAVIOR-1K~\cite{li2024} emphasizes complete household activities, combining navigation, bimanual manipulation, and dependent operations in house-scale environments across 1,000 human-centered tasks. OmniGibson supports articulated and deformable objects, liquids, and object-state changes, enabling diverse cooking, cleaning, and organizing activities. Predicate-based evaluation measures partial goal completion and full-task success, testing whether manipulation skills compose reliably over long horizons.
Challenge demonstrations include head and two wrist RGB-D views, proprioception, actions, camera poses, and language annotations \cite{behavior2026}. Raw HDF5 files additionally contain per-timestep simulator states and scene metadata, including robot and object states needed for trajectory replay and additional observation generation \cite{behavior2026,behaviorreplay2026}. Evaluation-time policy inputs remain restricted to onboard RGB, depth, and proprioception \cite{behavior2026}.

Compared with 2025, the 2026 challenge expands from 50 to 100 tasks across more scenes, retaining 200 demonstrations per task. Table~\ref{tab:benchmark_scale} compares dataset scale and evaluation protocols \cite{behavior2026,behavior2025}. Scene counts refer to distinct house-scale environment assets used by the challenge; the 2026 release adds four scenes to the original three. Its demonstrations~\cite{behavior2026} total approximately 1,950 hours.
Complementing full-task evaluation, Behavior-Skill~\cite{ma2026behaviorskill} tests 34 skill categories from valid intermediate states. ManiUnit~\cite{wei2026maniunit} measures skill sensitivity to starting-state variations through base-position and joint-configuration perturbations, and enables independent skill evaluation without executing preceding stages by restoring intermediate simulator states and defining local success conditions. Rasouli et al.~\cite{rasouli2026} measure object-handling safety and effects on surrounding objects.

\begin{table}[htbp]
\centering
\caption{Dataset scale and evaluation protocols of the 2025 and 2026 BEHAVIOR Challenges.}
\label{tab:benchmark_scale}
\begin{tabular}{lcc}
\toprule
Characteristic & 2025 & 2026 \\
\midrule
Household tasks & 50 & 100 \\
Human demonstrations per task & 200 & 200 \\
Mean human demonstration duration (s) & 397.04 & 351.54 \\
Distinct house-scale scenes & 3 & 7 \\
Skill annotation categories\textsuperscript{a} & 34 & 35 \\
Public instances per task used for reporting & 10 & 10 \\
Private test instances per task & 10 & 20 \\
Simulation timeout per task\textsuperscript{b} & $2.0\times\bar{T}_{\mathrm{task}}$ & $1.5\times\bar{T}_{\mathrm{task}}$ \\
\bottomrule
\end{tabular}
\par\smallskip
\begin{minipage}{\linewidth}
\footnotesize
\textsuperscript{a}Categories count distinct skill labels in the released annotations, including rare labels; the website summaries list 31 for both years.\\
\textsuperscript{b}$\bar{T}_{\mathrm{task}}$ is the mean human completion time for the particular task.\\
\end{minipage}
\end{table}

\subsection{Vision-Language-Action policies for BEHAVIOR}

Physical Intelligence's $\pi$ series~\cite{black2024,black2025} exemplifies generalist VLA policies that combine pretrained vision-language representations with continuous action generation. Both leading solutions in the 2025 BEHAVIOR Challenge build on $\pi_{0.5}$. The first-place Robot Learning Collective (RLC) solution~\cite{larchenko2025} uses stage prediction, action inpainting, temporal compression, and gripper correction, achieving a private-test Q-score of 0.260. The second-place Openpi Comet solution~\cite{bai2025} combines heterogeneous pretraining with iterative training on successful policy rollouts, achieving a private-test Q-score of 0.251. Galaxea G0.5 generates compact action codes in a single autoregressive stream. It achieves a Q-score of 0.314 on the 2025 BEHAVIOR validation set \cite{galaxea2026}. Rui et al.~\cite{rui2026} study agent-orchestrated composition of $\pi_{0.5}$ skills on BEHAVIOR, showing that completing one skill does not necessarily satisfy the next skill's execution preconditions.

\subsection{World action models}

World action models combine future visual prediction with action modeling. Within the Cosmos family, Cosmos Policy~\cite{kim2026cosmospolicy} adapts Cosmos-Predict2 to model actions, future states, and expected returns; its planning variant further refines the world model and value function using policy rollout data. Cosmos 3~\cite{nvidia2026} unifies language, images, video, audio, and actions within an omnimodal framework. Fast-WAM~\cite{yuan2026} retains video co-training but omits explicit future video generation at inference. Its controlled comparisons show that removing video co-training causes a larger performance drop than omitting test-time future generation, suggesting that predictive training can support effective action generation without explicit future imagination during execution. Cosmos Policy evaluates on LIBERO, RoboCasa, and real-world bimanual tasks, while Fast-WAM evaluates on LIBERO, RoboTwin 2.0, and real-world towel folding.

\subsection{Agentic Harnesses for BEHAVIOR}

Huang et al.'s RoboHarness~\cite{huang2026roboharness} uses Codex powered by GPT-5.5 to orchestrate heterogeneous policies through memory and execution feedback, with Memory Bridge supporting policy handoffs and online adaptation improving orchestration across episodes without joint policy retraining. On 6 selected BEHAVIOR-1K tasks, with 10 episodes per task, it combines $\pi_{0.5}$ with task-and-motion planning and achieves a mean progress score of 0.658 and a success rate of 43.3\%.
Qu et al.'s RoboHarness~\cite{qu2026roboharness} equips Qwen3.8-Flash-Next~\cite{qiu2026qwen38} with visual point selection, keypoint tracking, and geometric-constraint tools for robot control without model fine-tuning. On the first 10 tasks of the 2025 BEHAVIOR Challenge, with 5 instances per task, it achieves a mean Q-score of 0.445 using task-specific human-written prompts and execution budgets.
SenseNova et al.~\cite{joint2026astra} use demonstration-derived knowledge packages to support GPT-6-Astra's closed-loop embodied control, keeping prepared knowledge fixed during evaluation. In self-evaluation across all 100 tasks in the 2026 BEHAVIOR Challenge, with 1 instance per task, the system achieves a 37\% success rate under the official step budget. The combined cost of in-context preparation and standard-budget execution averages over 70 million tokens, including cached inputs, and over 400 minutes of wall-clock time per task. These costs highlight the expense of evaluating multiple instances and repeated trials.

\section{Method}
\label{sec:method}
\begin{figure}[H]
\centering
\reportfig[width=\textwidth]{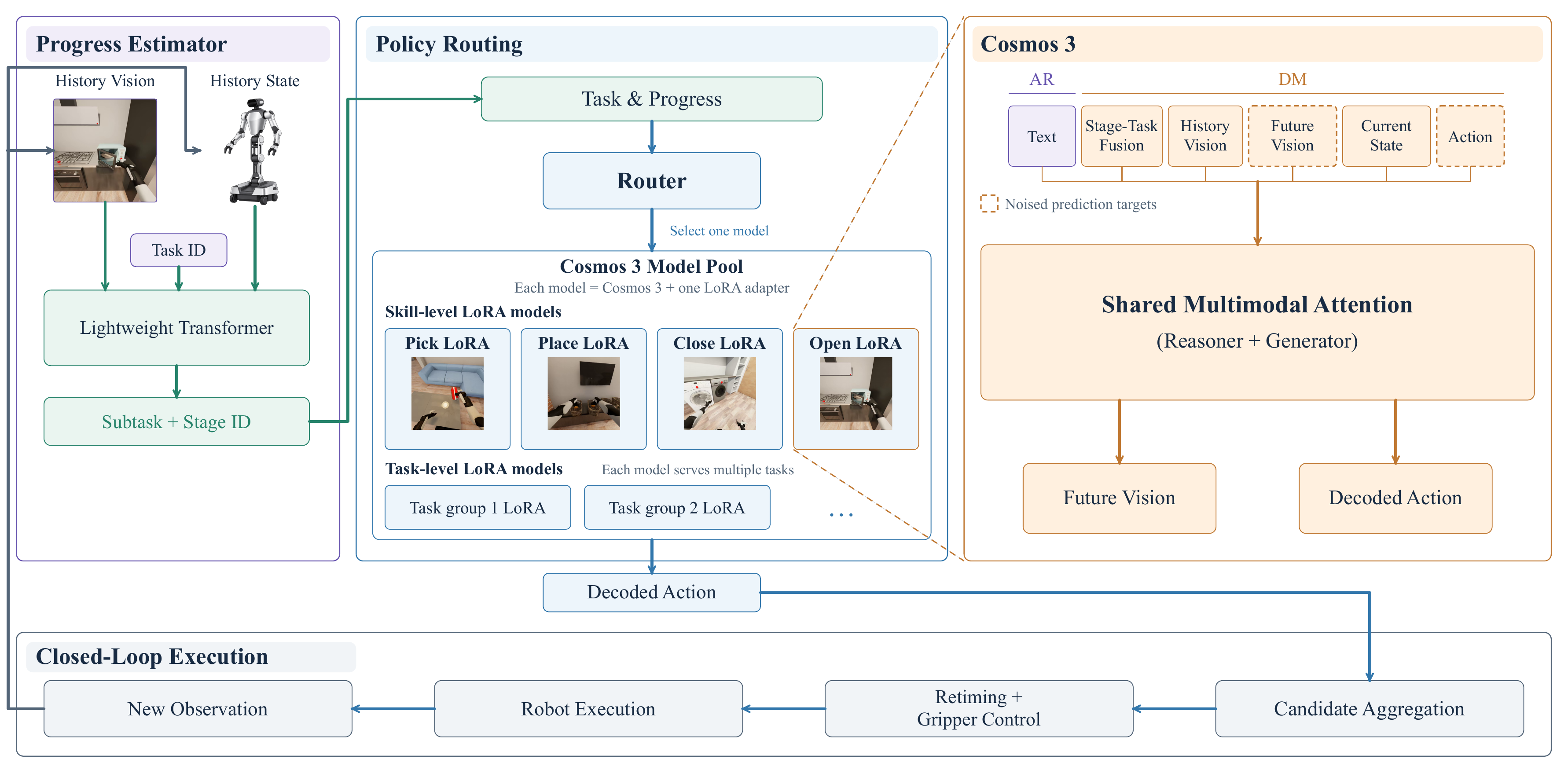}
\caption{Overall framework of our method. The progress estimator predicts the subtask and stage ID from visual and state histories and the task ID. These predictions guide skill-specific LoRA activation, overriding the task-configured base policy or task-group LoRA adapter. Cosmos 3 is conditioned on visual history, task and skill text, current state, and the 4 conditioning tokens produced by stage-task fusion. Generated action chunks undergo candidate aggregation, retiming, and gripper control for closed-loop execution.}
\label{fig:framework}
\end{figure}

{ 
Figure~\ref{fig:framework} provides an overview of our proposed framework. In this section, we present our method in three parts. First, Section~\ref{sec:cosmos_base} introduces Cosmos 3 as our base model. Next, Section~\ref{sec:ambiguity} details how we integrate richer visual history, task-progress estimation, and task-group LoRA fine-tuning to address observational ambiguity at complementary levels. Finally, Section~\ref{sec:inference} presents our inference and execution strategy. 
}

\subsection{Cosmos 3 as our base model}
\label{sec:cosmos_base}

\begin{figure}[!htbp]
\centering
\reportfig[width=\textwidth]{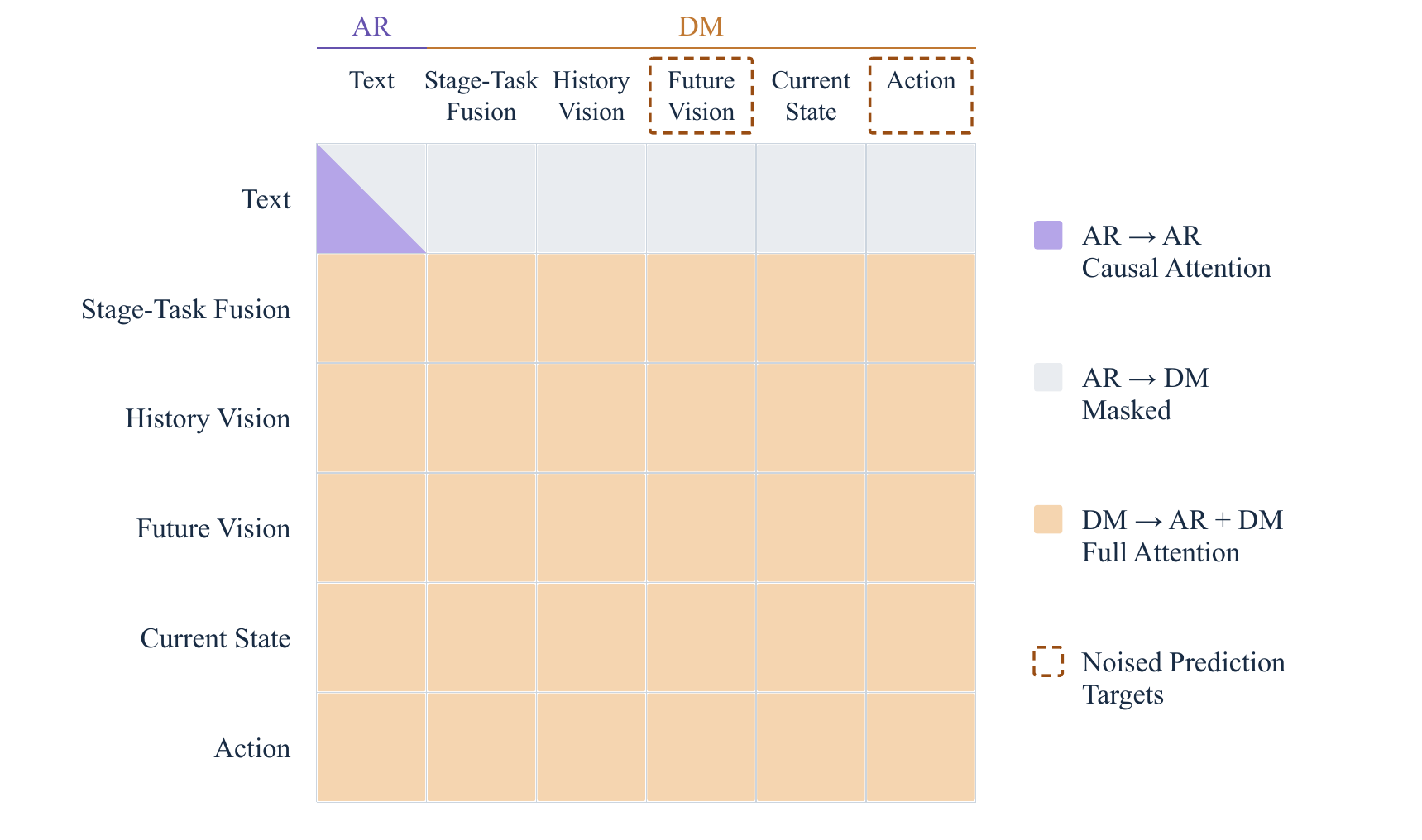}
\caption{Attention mask. Following Cosmos 3 \cite{nvidia2026}, AR attention is causal within AR, while DM attends to both subsequences. The model jointly denoises future vision and action tokens.}
\label{fig:architecture}
\end{figure}

Starting from Cosmos 3-Nano, a 16B-parameter version of Cosmos 3 \cite{nvidia2026}, we enhance it by integrating visual-history conditioning, explicit task-progress tokens, and skill text alongside the task instruction, and adopt state-relative delta actions following~\cite{larchenko2025}. We fine-tune the policy model on demonstrations from all 100 challenge tasks. Figure~\ref{fig:architecture} shows the attention mask.

\subsubsection{Policy architecture}
\label{sec:representation}

Stage-task fusion encodes the task embedding and stage ID as 4 clean conditioning tokens. We extend the Cosmos 3 input sequence by inserting these tokens, together with clean tokens for visual history and current proprioception, into the diffusion (DM) subsequence. We retain the Cosmos 3 attention mask \cite{nvidia2026}: the reasoner attends causally within the autoregressive (AR) subsequence, while the generator attends to all AR and DM tokens. The added conditioning tokens are therefore visible only to the generator.

\begin{figure}[htbp]
\centering
\reportfig[width=\textwidth]{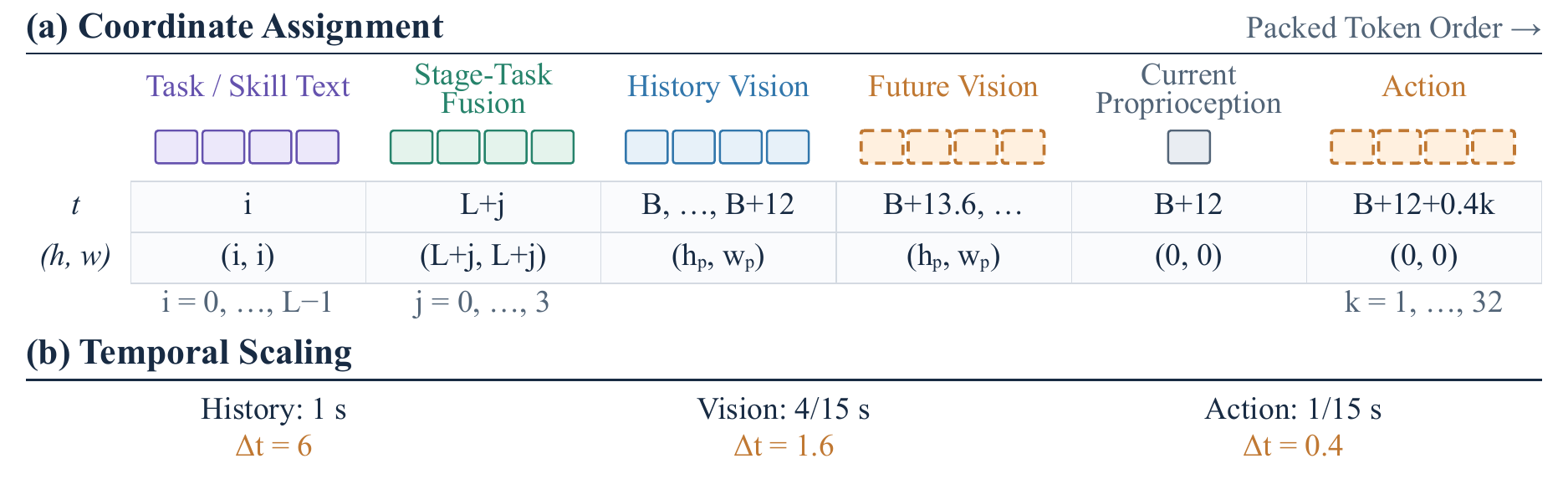}
\caption{Multimodal position encoding. (a) Packed token groups and representative coordinates for our policy, with the current observation at $\tau=2$~s. Group sizes are schematic; visual tokens retain patch coordinates $(h_p,w_p)$, and dashed outlines denote prediction targets. Token order does not imply temporal order: current proprioception remains aligned with the current image. (b) Elapsed time maps to temporal-coordinate increments at 6 units per second.}
\label{fig:coordinate_assignment}
\end{figure}

Cosmos 3's 3D multimodal rotary position encoding (mRoPE) assigns coordinates $(t,h,w)$ to each token \cite{nvidia2026}. Text tokens and the tokens produced by stage-task fusion use increasing coordinates with $t=h=w$; vision retains spatial patch coordinates, whereas proprioception and actions use $h=w=0$. Vision, proprioception, and actions share the temporal mapping
\begin{equation}
t=B+\omega\tau,\qquad \omega=\frac{f_{\mathrm{ref}}}{c}=6\,\mathrm{s}^{-1},
\qquad B=L+4+15{,}000,
\label{eq:temporal_coordinates}
\end{equation}
where $t$ is a dimensionless temporal coordinate and $\tau$ is the elapsed time in seconds from the earliest available conditioning observation. The offset $B$ accounts for the packed text length $L$ (including EOS and generation-start tokens), 4 positions for the stage-task fusion output tokens, and a modality margin of 15,000. Here, $f_{\mathrm{ref}}=24$~FPS is the Cosmos 3 reference frame rate, and $c=4$ is the Wan2.2 VAE's temporal compression factor~\cite{wan2025wan22}, corresponding to 4 input-frame intervals per latent time step. These values give $\omega=6\,\mathrm{s}^{-1}$, independently of our 15~Hz training grid. Thus, the 3 conditioning frames spanning 2~s in Figure~\ref{fig:coordinate_assignment}(a) have temporal coordinates $B$, $B+6$, and $B+12$. Current proprioception shares the temporal coordinate of the current observation, while future actions advance by $6/15=0.4$ coordinate units per step.

The 23-dimensional current proprioceptive state, including base velocity, is normalized to $[-1,1]$ using global 1st- and 99th-percentile statistics. It serves as clean conditioning and is excluded from the action loss.

Following Larchenko et al.\ \cite{larchenko2025}, we express selected joint targets relative to the chunk-start state $x^{\mathrm{state}}_d$. For command $u_{i,d}$ at chunk position $i$ and dimension $d$, the training target is
\begin{equation}
b_{i,d} =
\begin{cases}
u_{i,d}-x^{\mathrm{state}}_d, & d\in\mathcal D_{\Delta}, \\
u_{i,d}, & d\notin\mathcal D_{\Delta},
\end{cases}
\qquad i=0,\ldots,H-1.
\label{eq:delta_actions}
\end{equation}
Here, $H=32$ and $\mathcal D_{\Delta}$ comprises the first 3 torso joints and 7 joints per arm (17 dimensions). The 3 base velocities, 4th torso joint, and 2 gripper commands remain unchanged.

We normalize all 23 dimensions using quantiles specific to each chunk position and dimension:
\begin{equation}
\bar a_{i,d}=\operatorname{clip}\!\left(
2\frac{b_{i,d}-Q_{0.01,i,d}}{Q_{0.99,i,d}-Q_{0.01,i,d}+\epsilon}-1,\;-1,\;1\right),
\qquad \epsilon=10^{-6}.
\label{eq:action_normalization}
\end{equation}
Here, $\bar a_{i,d}$ is the normalized model-space target, and $Q_{p,i,d}$ is the $p$-quantile of $b_{i,d}$ across training windows. These action statistics are estimated separately from the global proprioceptive-state statistics.

At inference, we reverse the affine normalization and add $x^{\mathrm{state}}_d$ back only to dimensions in $\mathcal D_{\Delta}$ to recover environment commands.

\subsubsection{Implementation details}
\label{sec:data}

Our training set contains 20,000 demonstrations across 100 challenge tasks, with 200 demonstrations per task. Sampling on the 15~Hz training grid yields 104,823,365 eligible windows. We configure a 600,000-step training schedule for the base model. It consumes approximately 7,095 GPU-hours, with an effective batch size of 384. This schedule corresponds to approximately 2.20 equivalent epochs over the whole window pool. We use fused AdamW to update only generator-specific weights, time embeddings, modality projections, and task--stage conditioning, with no warmup, linear decay from $8\times10^{-5}$ to zero, and a $5\times$ learning-rate multiplier for action and task--stage modules. Our training budget is constrained by our computing resources, and we expect longer training to yield further improvements.

We resize the head view to $224\times224$ pixels and each wrist view to $112\times112$ pixels, and place the head view above the side-by-side left and right wrist views to form a $336\times224$ (height $\times$ width) RGB image. The policy uses Cosmos 3's 480p preprocessing configuration and predicts 32-step chunks of 23-dimensional actions on a 15~Hz grid. To avoid repeated visual encoding during training, we precompute visual latents with the Wan2.2 VAE and store them per window in packed 6-bit form with groups of 64 values. With this cached-latent pipeline and bfloat16 training, the mean model FLOPs utilization (MFU) is 52.3\%. At deployment, the policy encodes live observations without this cache quantization.

The original single-frame conditioning recipe averages the visual rectified-flow loss over 1 clean and 8 predicted visual latents, including zero-error clean tokens in the denominator. Its nominal weight of 10 thus gives an effective prediction-loss weight of $10\times8/9$, which would decrease with additional clean history latents. We instead average over prediction tokens only and use a weight of $80/9\approx8.889$, preserving the original effective weight regardless of history length.

\subsection{Addressing observational ambiguity}
\label{sec:ambiguity}

Long-horizon household tasks exhibit non-Markovian ambiguity: visually similar observations can occur at different task stages, so the current observation alone may be insufficient to determine the next action \cite{larchenko2025}. We address this ambiguity by incorporating visual history and explicit task-progress conditioning. Task-group LoRA fine-tuning, together with task and skill text, further distinguishes task-dependent behaviors. Skill-specific LoRA adapters specialize manipulation operations for execution guided by the progress estimator.

\subsubsection{Visual history}
\label{sec:history}

Our policy conditions on the current observation and 2 historical observations spanning 2~s. These observations are encoded as clean visual latents sampled at 1~s intervals, with temporal positions assigned from their actual timestamps.

\subsubsection{Task-progress conditioning and skill text}
\label{sec:classifier}
\label{sec:prompts}

Following Larchenko et al.\ \cite{larchenko2025}, we use task-specific stage counts and divide each demonstration into equal-duration stages. For task $q$, let $\bar N_q$ be its mean demonstration length, $N_e$ the length of demonstration $e$, and $n$ its zero-based chunk-start frame index, all measured on the 30~Hz source timeline. The stage count $K_q$ and ground-truth stage ID $s$ are defined as follows:
\begin{equation}
K_q=\operatorname{clip}\!\left(\left\lceil\frac{\bar N_q}{900}\right\rceil,5,15\right),
\qquad
s=\operatorname{clip}\!\left(\left\lfloor\frac{K_q n}{N_e}\right\rfloor,0,K_q-1\right).
\label{eq:stage_labels}
\end{equation}
The 900-frame reference duration corresponds to 30~s and determines $K_q$; each demonstration is then divided into $K_q$ equal stages of length $N_e/K_q$ frames. Across the 100 tasks, there are 1,120 task-specific stages. These labels encode temporal progress rather than annotated skill boundaries.

For task $q$ with $K_q$ stages, let $\mathbf e_q\in\mathbb R^{d_{\mathrm{model}}}$ be its learned task embedding, where $d_{\mathrm{model}}$ is the DM token dimension. Stage $s$ is represented by 2 features of dimension $d_{\mathrm{model}}/2$: a learned task--stage embedding $\mathbf v_{q,s}$ and $\mathbf p=\mathrm{PE}(s/(K_q-1))$. Here, $\mathrm{PE}$ concatenates sine and cosine features with logarithmically spaced periods from $10^{-3}$ to $1$. Their concatenation $\mathbf x=[\mathbf e_q;\mathbf p;\mathbf v_{q,s}]\in\mathbb R^{2d_{\mathrm{model}}}$ produces gates $\mathbf g_r=\operatorname{sigmoid}(W_r\mathbf x+\mathbf b_r)$ for $r\in\{e,p,v\}$, each matching its corresponding feature's dimension. The 4 tokens $\mathbf z_i\in\mathbb R^{d_{\mathrm{model}}}$ are defined as
\begin{equation}
\begin{aligned}
\mathbf z_1 &= \mathbf e_q\odot\mathbf g_e,\\
\mathbf z_2 &= W_2\,\mathrm{ReLU}(W_1\mathbf x+\mathbf b_1)+\mathbf b_2,\\
\mathbf z_3 &= W_{\mathrm{stage}}[\mathbf p\odot\mathbf g_p;\mathbf v_{q,s}\odot\mathbf g_v]+\mathbf b_{\mathrm{stage}},\\
\mathbf z_4 &= [\mathbf p;\mathbf v_{q,s}],
\end{aligned}
\label{eq:stage_tokens}
\end{equation}
where $W$ and $\mathbf b$ denote learned weights and biases shared across tasks and stages. The 4 tokens encode gated task features, joint task--stage fusion, gated stage features, and ungated stage features, respectively. We retain these 4 tokens following Larchenko et al.\ \cite{larchenko2025}, omit the additional base-task token, and use them as clean conditioning inputs in the DM subsequence as shown in Figure~\ref{fig:architecture}.

At inference, a separate stage-estimation branch supplies task-progress predictions; its architecture, post-processing, and skill-to-stage alternative are described in Section~\ref{sec:planner}.

Text conditioning combines the task instruction with the current-skill instruction, appended with probability 0.5 when skill annotations are available. Independently, caption dropout removes the entire prompt with probability 0.1.

\subsubsection{Task-group LoRA fine-tuning}
\label{sec:adapters}

To distinguish tasks that have similar initial observations but different required behaviors, we divide the tasks into groups according to the scene and position metadata. For each task, we identify the scene and the robot's initial room from its initial planar positions across 200 demonstrations. Tasks that share the same scene--room pair are connected in a conflict graph, and balanced graph coloring assigns them to different groups. The largest clique consists of 17 tasks that start in the same kitchen, so at least 17 groups are required. Our partition meets this lower bound, yielding 15 groups of 6 tasks and 2 groups of 5. Table~\ref{tab:groups} lists the task assignment, and Figure~\ref{fig:groups} in Appendix~\ref{app:views} shows their initial observations.

\begin{table}[H]
\centering
\caption{Task IDs assigned to each group.}
\label{tab:groups}
\begin{tabular}{rl@{\hspace{2em}}rl}
\toprule
Group & Tasks & Group & Tasks \\
\midrule
1 & 1, 4, 37, 44, 83, 85 & 10 & 14, 31, 45, 52, 79, 89 \\
2 & 3, 5, 10, 59, 68, 86 & 11 & 16, 32, 46, 61, 87, 92 \\
3 & 0, 7, 8, 11, 69, 90 & 12 & 33, 47, 57, 70, 88, 93 \\
4 & 2, 9, 12, 21, 25, 72 & 13 & 6, 23, 48, 58, 65, 94 \\
5 & 17, 22, 26, 34, 54, 81 & 14 & 15, 50, 53, 71, 75, 97 \\
6 & 20, 28, 30, 40, 74, 91 & 15 & 36, 55, 62, 76, 77, 98 \\
7 & 18, 24, 29, 42, 56, 95 & 16 & 38, 64, 78, 84, 99 \\
8 & 27, 41, 49, 60, 63, 96 & 17 & 35, 39, 67, 80, 82 \\
9 & 13, 19, 43, 51, 66, 73 & & \\
\bottomrule
\end{tabular}
\end{table}

We additionally use successful rollout data from Openpi Comet~\cite{bai2025} for LoRA fine-tuning. For each group, we train a LoRA adapter \cite{hu2022} with rank 128 and $\alpha=256$, based on our base policy model. The adapters are applied to the action generator's attention projections (query, key, value, and output) and MLP projections (gate, up, and down). Conditioning and prompt sampling follow the setup described above. We train each LoRA adapter for 40,000 steps. We warm up the learning rate for 500 steps to $2\times10^{-4}$, then decay it linearly to zero. We use a weight decay of 0.05. Training the 17 adapters requires approximately 6,427 GPU-hours in total, using an effective batch size of 384. We use the 40,000-step checkpoint at deployment.

\subsubsection{Skill-specific LoRA adapters}
\label{sec:skilladapters}

To improve the reliability of individual manipulation operations, we pool demonstration segments of the same skill across tasks and train separate supervised LoRA adapters for picking, placing, opening, and closing. These adapters use the same LoRA and optimizer configurations as task-group LoRA fine-tuning. For picking, opening, and closing, training intervals are extended backward to include immediately preceding sequences of \emph{push to} and \emph{turn to}, stopping at navigation or unrelated skills. Task-specific rules additionally include selected \emph{hand over} or \emph{tip over} preparations. Placement segments retain their original annotated boundaries. Training windows are selected by their start frames, with twice the sampling weight in the latter half of each selected interval.

Table~\ref{tab:skills} lists the adapters and their corresponding skill categories. At test time, the router selects and activates the appropriate skill-specific LoRA adapter using skill predictions from the progress estimator (Section~\ref{sec:planner}). Only the skill categories listed in Table~\ref{tab:skills} trigger the corresponding adapter; preparatory actions included in the extended training segments do not. Appendix~\ref{app:skillviews} provides panels for visual examples of the 4 skill families.

\begin{table}[H]
\centering
\caption{Skill-specific LoRA adapters and corresponding skill categories.}
\label{tab:skills}
\begin{tabular}{lp{0.72\textwidth}}
\toprule
LoRA adapter & Skill category \\
\midrule
Pick & pick up from \\
Place & place in, place in next to, place on, place on next to, place under \\
Close & close door, close lid, close drawer \\
Open & open door, open lid, open drawer \\
\bottomrule
\end{tabular}
\end{table}

\subsection{Inference and execution}
\label{sec:inference}

In the inference stage, the progress estimator provides task-progress context and skill predictions, which guide the router in selecting policies and skill-specific LoRA adapters. Action-candidate aggregation, rolling inpainting, and temporal and gripper control support execution.

\subsubsection{Progress Estimator}
\label{sec:planner}

\begin{figure}[!htbp]
\centering
\reportfig[width=\textwidth]{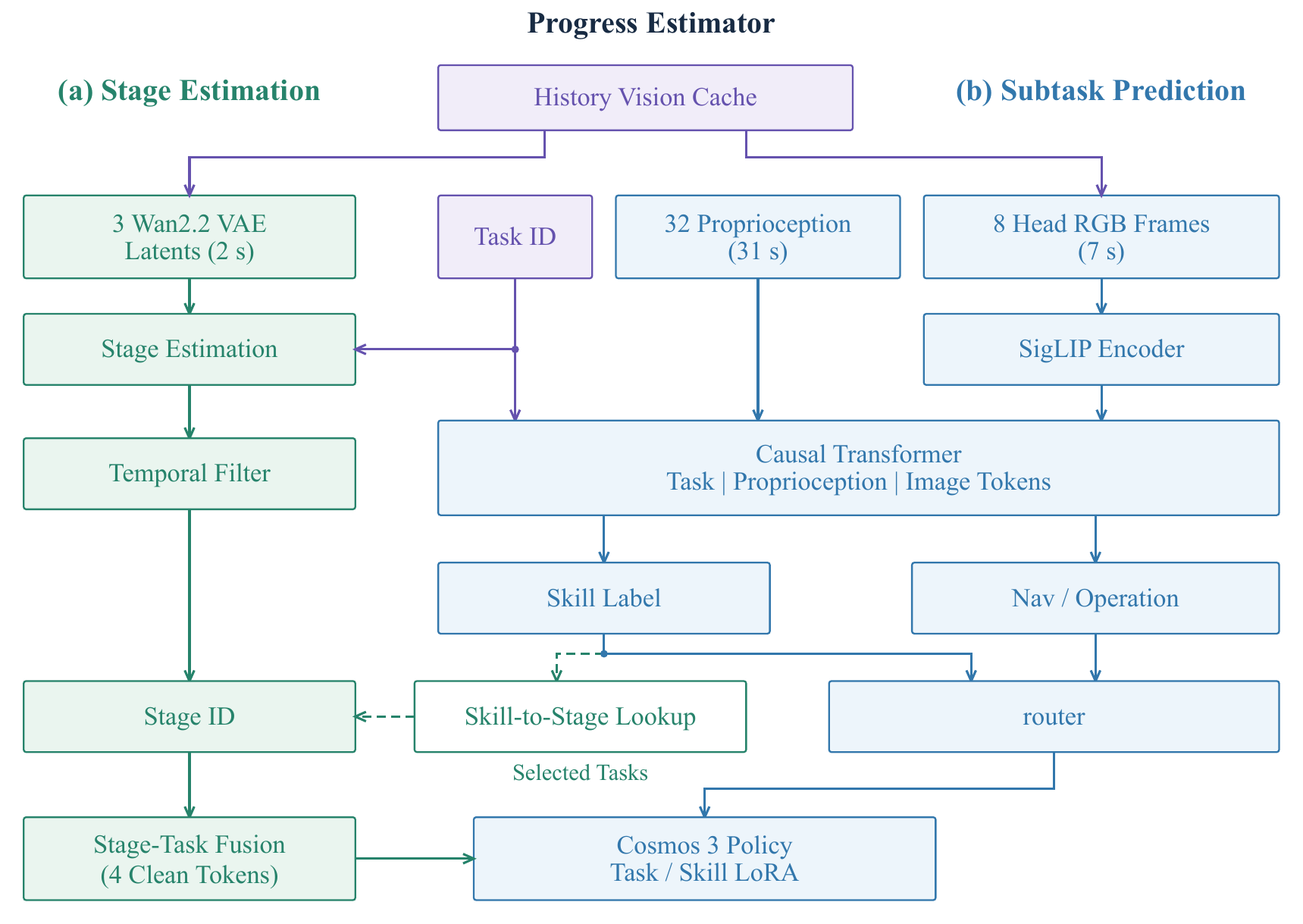}
\caption{Progress estimator for stage conditioning and policy routing. The stage-estimation and subtask-prediction branches share a visual-history cache and task ID. Skill predictions and navigation/operation status guide the router; the dashed skill-to-stage lookup supplies alternative stage IDs for selected tasks.}
\label{fig:stage_planner}
\end{figure}

The progress estimator combines independently trained stage-estimation and subtask-prediction branches to provide task-progress context and guide policy routing (Figure~\ref{fig:stage_planner}). Both use a shared visual-history cache and task ID. The subtask-prediction architecture could also be extended to jointly predict subtasks and stage IDs, but competition time constraints prevented us from evaluating this alternative.

The stage-estimation branch processes 3 Wan2.2 VAE latents spaced 1~s apart and a task embedding using a 2-layer Transformer (4096 hidden dimensions, 32 heads; approximately 404M parameters excluding the shared VAE). With mRoPE and mean pooling, it predicts 15 stage logits masked to valid task stages and is trained by cross-entropy on the labels in Eq.~\eqref{eq:stage_labels}. Task-specific post-processing (Appendix~\ref{app:stage_rules}) stabilizes predictions before encoding them as 4 clean conditioning tokens.

The subtask-prediction branch takes 32 proprioceptive states (61 dimensions), 8 head-camera images, both sampled at 1~Hz, and the task ID. The action policy derives its 23-dimensional state from this input using base velocities, torso and arm joint positions, and one opening value per gripper. The subtask-prediction model combines a fine-tuned SigLIP-B/16 encoder~\cite{zhai2023siglip} with an 8-layer causal Transformer (128 hidden dimensions, 4 heads), totaling 94.6M parameters. The subtask-prediction branch predicts whether the robot is navigating or manipulating and identifies the current subtask by its skill and associated objects. Navigation/operation predictions gate entry into skill branches, while skill predictions determine which branch to activate. Skill-specific LoRA adapters take precedence over the task-configured base policy or task-group LoRA adapter.

For selected tasks, a skill-to-stage lookup derived from training statistics maps predicted skill labels to stage IDs, replacing direct stage estimation and temporal filtering. Stage IDs provide conditioning, while the router selects adapters using skill predictions and navigation/operation status.

To examine stage-conditioning sensitivity, we vary the input stage ID over 2,000 demonstration windows from task 26 ($K_q=15$), holding the model, observations, and per-window noise seeds fixed. Let $\mathcal W_s$ denote windows with ground-truth stage $s$, and $s'$ the input stage, with $s,s'\in\{0,\ldots,K_q-1\}$. Using the action notation of Eq.~\eqref{eq:delta_actions}, $u^{(\ell)}_{i,d}$ and $u^{(\ell,s')}_{i,d}$ denote the demonstrated command and decoded single-candidate prediction conditioned on $s'$ in window $\ell$, respectively. With $D=23$ action dimensions, we compute
\begin{equation}
\begin{aligned}
E_{s,s'}
&=\frac{1}{|\mathcal W_s|HD}
\sum_{\ell\in\mathcal W_s}\sum_{i=0}^{H-1}\sum_{d=1}^{D}
\left|u^{(\ell,s')}_{i,d}-u^{(\ell)}_{i,d}\right|,\\
\Delta^{\mathrm{MAE}}_{s,s'}
&=\frac{E_{s,s'}-E_{s,s}}{E_{s,s}}\times100\%.
\end{aligned}
\label{eq:stage_mae_change}
\end{equation}
Here, $E_{s,s}$ is the ground-truth-conditioned baseline on the same windows. Positive heatmap values indicate increased error.

\begin{figure}[!htbp]
\centering
\reportfig[width=0.95\textwidth]{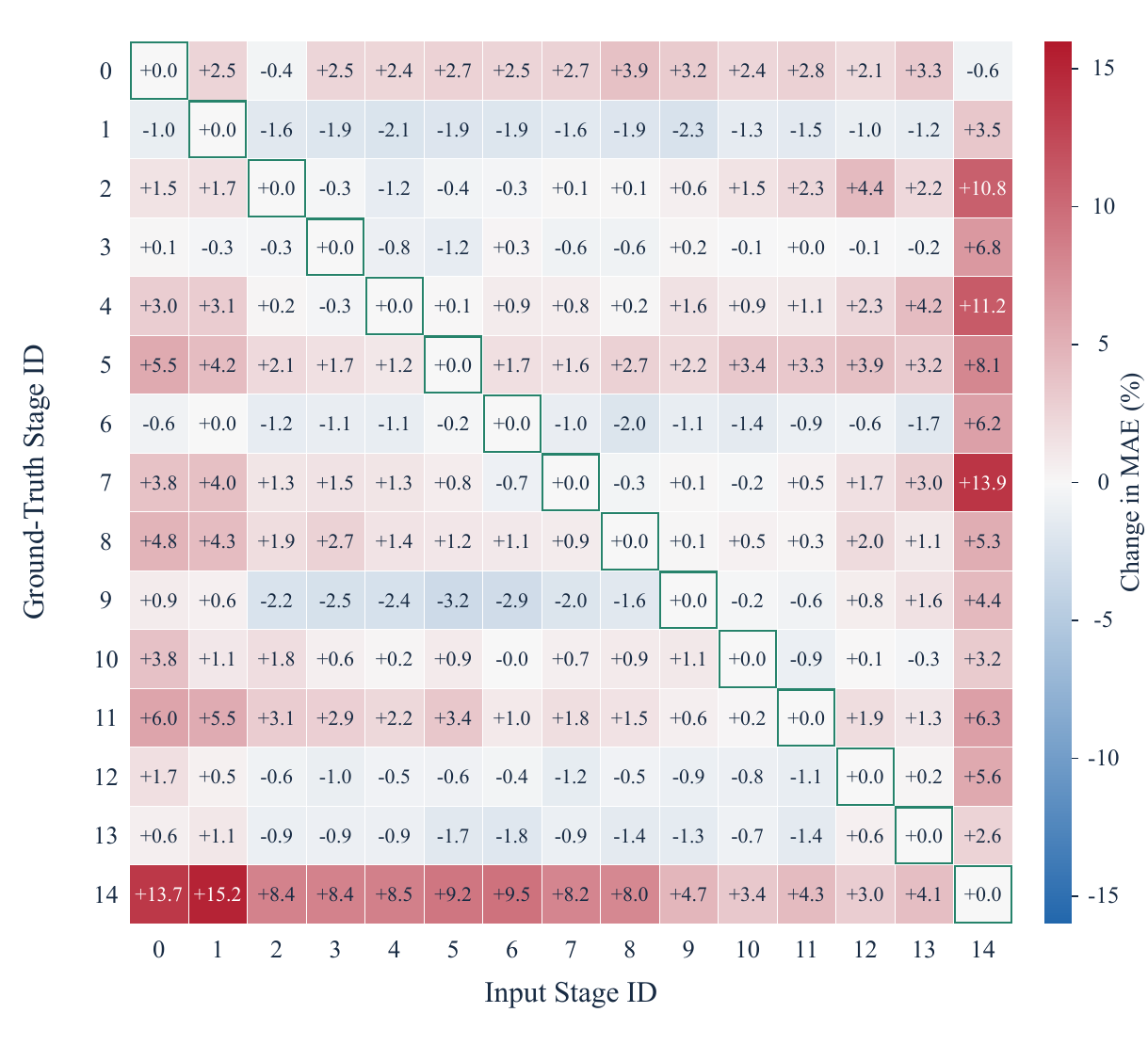}
\caption{Stage-ID sensitivity on task 26. Each cell shows the percentage change in mean action-prediction MAE relative to using the ground-truth stage ID on the same windows. Rows group windows by ground-truth stage; columns specify the input stage. Positive values indicate higher error.}
\label{fig:stage_sensitivity}
\end{figure}

Figure~\ref{fig:stage_sensitivity} shows that sensitivity to the input stage ID varies across task stages and is particularly pronounced at the final stage ($s=14$).

\subsubsection{Action-candidate aggregation and rolling inpainting}
\label{sec:sampling}

We generate $N=5$ candidates by jointly denoising future visual latents and actions from independent Gaussian noise. At step $k$, let $\mathbf v_k^{\mathrm{cond}}$ and $\mathbf v_k^{\mathrm{uncond}}$ be the velocity predictions with the text prompt and an empty prompt, respectively; both retain the same visual, proprioceptive, and task--stage conditions. Classifier-free guidance \cite{ho2022} uses
\begin{equation}
\mathbf v_k^{\mathrm{CFG}}=\mathbf v_k^{\mathrm{uncond}}+
\gamma\big(\mathbf v_k^{\mathrm{cond}}-\mathbf v_k^{\mathrm{uncond}}\big),
\qquad \gamma=3.
\label{eq:inference_cfg}
\end{equation}
Following the shift parameterization of Cosmos 3 \cite{nvidia2026}, we use UniPC \cite{zhao2023unipc} with $S=4$ denoising steps and shift factor $\beta=5$:
\begin{equation}
\tilde\sigma_k=\tilde\sigma_0\left(1-\frac{k}{S}\right),\qquad
\sigma_k=\frac{\beta \tilde\sigma_k}{1+(\beta-1)\tilde\sigma_k},\qquad
k=0,\ldots,S-1,\qquad \sigma_S=0.
\label{eq:inference_shift}
\end{equation}
Here, $k$ indexes denoising steps, $\tilde\sigma_k$ is the unshifted noise level, and $\sigma_k$ is the shifted noise level used by UniPC. With $\tilde\sigma_0=0.999$, the 4 noise levels are
\[
(\sigma_0,\sigma_1,\sigma_2,\sigma_3)\approx(0.999800,\;0.937265,\;0.833055,\;0.624687),
\]
followed by the noise-free endpoint $\sigma_4=0$. Starting from $\mathbf y_0\sim\mathcal N(0,I)$, the noisy sample $\mathbf y_k$ gives the clean-sample estimate
\begin{equation}
\widehat{\mathbf y}^{\mathrm{clean}}_k=\mathbf y_k-\sigma_k\mathbf v_k^{\mathrm{CFG}},
\qquad k=0,\ldots,3.
\label{eq:four_step_decoding}
\end{equation}
Second-order UniPC uses these estimates to advance between noise levels, with rolling inpainting applied after each step as defined below. The final action predictions are decoded to physical commands using Section~\ref{sec:representation} before aggregation.

Let $U^{(j)}=[u^{(j)}_{i,d}]\in\mathbb R^{H\times D}$ denote candidate $j$ in physical action coordinates, consistent with $u_{i,d}$ in Eq.~\eqref{eq:delta_actions}. Here, $H=32$, $D=23$, $i=0,\ldots,H-1$, and $j=1,\ldots,N$. The aggregated chunk is $\hat U=[\hat u_{i,d}]$. Partition the dimensions into grippers $\mathcal G$ and continuous commands $\mathcal C$. For each gripper, define $o^{(j)}_{i,d}=\mathbf 1[u^{(j)}_{i,d}\geq0]$ and let $\mathcal I_{i,d}$ contain candidates whose indicator matches the majority. Mean/vote aggregation is
\begin{equation}
\hat u_{i,d}=
\begin{cases}
\displaystyle\frac{1}{N}\sum_{j=1}^{N}u^{(j)}_{i,d}, & d\in\mathcal C,\\[6pt]
\displaystyle\frac{1}{|\mathcal I_{i,d}|}\sum_{j\in\mathcal I_{i,d}}u^{(j)}_{i,d}, & d\in\mathcal G.
\end{cases}
\label{eq:action_mean_vote}
\end{equation}

Alternatively, medoid aggregation selects a complete candidate using pairwise distances $\Delta_{jm}$ between candidates $j$ and $m$, combining normalized continuous-action differences and gripper sign disagreement:
\begin{equation}
\begin{aligned}
\Delta_{jm}&=\left[\frac{1}{HD}\sum_{i=0}^{H-1}\left(
\sum_{d\in\mathcal C}\left(\frac{u^{(j)}_{i,d}-u^{(m)}_{i,d}}{\rho_{i,d}}\right)^2
+\sum_{d\in\mathcal G}\mathbf 1[o^{(j)}_{i,d}\ne o^{(m)}_{i,d}]
\right)\right]^{1/2},\\
j^\star&=\operatorname*{arg\,min}_{j\in\{1,\ldots,N\}}\sum_{m=1}^{N}\Delta_{jm},
\qquad \hat U=U^{(j^\star)}.
\end{aligned}
\label{eq:action_medoid}
\end{equation}
Here, $\rho_{i,d}=(Q_{0.99,i,d}-Q_{0.01,i,d}+\epsilon)/2$ is the physical change corresponding to one normalized action unit, using the active policy's normalization statistics and the $\epsilon$ defined in Eq.~\eqref{eq:action_normalization}.

Rolling inpainting retains an overlap from the preceding aggregated chunk $\hat U$. The executed source window determines this overlap (Section~\ref{sec:execution}), which is re-encoded relative to the new observed state and normalized using Eqs.~\eqref{eq:delta_actions}--\eqref{eq:action_normalization} to form the shared anchor $\bar{\mathbf h}$. At post-step noise level $\sigma=\sigma_{k+1}$, let $\bar{\mathbf a}^{(j)}_{\sigma,\mathrm{pre}}$ and $\bar{\mathbf a}^{(j)}_{\sigma,\mathrm{post}}$ denote candidate $j$'s overlap before and after inpainting, both in the normalized coordinates of Eq.~\eqref{eq:action_normalization}. With fixed initial Gaussian noise $\boldsymbol\xi^{(j)}$ on this overlap, the update is
\begin{equation}
\bar{\mathbf a}^{(j)}_{\sigma,\mathrm{post}}=
\begin{cases}
\frac{1}{2}\big[(1-\sigma)\bar{\mathbf h}+\sigma\boldsymbol\xi^{(j)}\big]
+\frac{1}{2}\bar{\mathbf a}^{(j)}_{\sigma,\mathrm{pre}}, & \sigma>0.7,\\[3pt]
\bar{\mathbf a}^{(j)}_{\sigma,\mathrm{pre}}, & \sigma\leq0.7.
\end{cases}
\label{eq:rolling_inpainting}
\end{equation}
The first prediction has no anchor. After aggregation, action chunks are upsampled to the simulator control grid using cubic interpolation for continuous commands and zero-order hold for grippers.

\subsubsection{Temporal retiming and execution windows}
\label{sec:execution}

We upsample each aggregated 32-step, 15~Hz action chunk into 64 commands for execution at \(f_{\mathrm{sim}}=30\)~Hz. To accelerate execution, we resample the first 26 commands into 20 (short-window) or the first 52 into 40 (long-window), scaling base velocity accordingly. Excessive gripper variation triggers an uncompressed prefix instead. Inpainting anchors follow the executed window.

We use TOPP-RA \cite{pham2018} as an alternative to uniform resampling to minimize traversal time along the predicted path subject to base and joint velocity limits. TOPP-RA uses the first 52 commands of the upsampled chunk as its source window. Integrating the base velocities and retaining the 18 joint targets yields a 21-dimensional path \(\mathbf q(\zeta)\), initialized at the measured position and velocity, with source-time coordinate \(\zeta\in[0,52/f_{\mathrm{sim}}]\). Grippers are handled separately. Let \(\chi=\dot\zeta^2\), \(\alpha=\ddot\zeta\), and \(\delta_r=\zeta_{r+1}-\zeta_r\) on the solver grid. Dots denote derivatives with respect to execution time and primes with respect to \(\zeta\). The chain rule gives the first 2 relations below; assuming constant path acceleration within each grid interval yields the discrete update:
\begin{equation}
\dot{\mathbf q}=\mathbf q'(\zeta)\sqrt{\chi},\qquad
\ddot{\mathbf q}=\mathbf q''(\zeta)\chi+\mathbf q'(\zeta)\alpha,\qquad
\chi_{r+1}=\chi_r+2\delta_r\alpha_r.
\label{eq:toppra_dynamics}
\end{equation}
Simulator limits constrain body-frame base and joint velocities:
\begin{equation}
\left|\mathcal R(\mathbf q(\zeta_r))\mathbf q'(\zeta_r)\right|\sqrt{\chi_r}
\leq \mathbf v_{\mathrm{lim}},\qquad \chi_r\geq0,
\label{eq:toppra_velocity}
\end{equation}
where \(\mathcal R\) rotates planar translation into the base frame without changing yaw or joint velocities, and \(\mathbf v_{\mathrm{lim}}\) contains the corresponding simulator velocity limits. The inequality is componentwise. Without torque or explicit acceleration bounds, this formulation does not guarantee dynamic feasibility or tracking accuracy.

A backward pass finds controllable speed intervals; a forward pass selects the fastest admissible progression from the measured entry velocity. For \(M\) grid intervals, the resulting squared path speeds \(\chi_r^\star\) determine the minimum traversal time and requested execution length \(H_{\mathrm{exec}}\):
\begin{equation}
T_{\min}=\sum_{r=0}^{M-1}\frac{2\delta_r}{\sqrt{\chi_r^\star}+\sqrt{\chi_{r+1}^\star}},\qquad
H_{\mathrm{exec}}=\max\!\left(20,\;10\left\lceil\frac{f_{\mathrm{sim}}T_{\min}}{10}\right\rceil\right).
\label{eq:toppra_duration}
\end{equation}
We accept $H_{\mathrm{exec}}\in\{20,30,40,50\}$ commands, with a 20-command minimum. A fixed-duration solve fits the path to $H_{\mathrm{exec}}/f_{\mathrm{sim}}$ seconds while preserving the measured entry velocity and predicted source-path exit velocity. Retimed poses determine base commands, joints follow the path, and grippers use zero-order hold. Trajectories must pass velocity, joint-position, and discrete path-reconstruction checks, reach the source endpoint, and retain matching inpainting anchors.

If $H_{\mathrm{exec}}>50$, we use linear 52-to-50 resampling; other rejections use uniform 52-to-40 resampling. The gripper-variation bypass takes precedence and executes the first 40 commands uncompressed. These fallbacks do not inherit TOPP-RA guarantees.

Following Larchenko et al.~\cite{larchenko2025}, we apply demonstration-derived, task- and stage-specific gripper rules. These rules override closing commands when the gripper should remain open, helping recovery from premature closure.

\section{Experiments}

Competition time and compute constraints limited strictly controlled ablations, and some comparisons cover only selected task subsets. OmniGibson was also updated during the competition, so the reported results span versions from v3.9.1 to v3.9.3.

We evaluate our system on all 100 tasks of the 2026 BEHAVIOR Challenge, using 10 public instances per task and a simulation-time limit of 1.5 times the task's mean human completion time. Predicate-based Q-score and success rate measure partial goal completion and full-task completion, respectively. Tables use Val for self-evaluation on public validation instances, Test for official private-test results, and SR for success rate.

Comparisons use matched task--instance sets, with task coverage reported in each table. Restricting evaluation to tasks with complete results may introduce selection bias. Appendix~\ref{app:per_task} provides per-task results.

We first report overall system performance and route selection, followed by full-suite comparisons of task-group LoRA fine-tuning, stage estimation, skill-specific routing, and execution strategies.

\subsection{Overall results}
\label{sec:results_routing}

Table~\ref{tab:routed} reports the top 6 official 2026 results, alongside 2025 baselines and our self-evaluation results for all tasks and the 2025 (IDs 0--49) and new 2026 (IDs 50--99) subsets. The official 2026 ranking uses the mean private-test Q-score across all 100 tasks. Differences in instance coverage and evaluation protocols limit comparisons with the 2025 baselines. Our system fixes policy routes before evaluation; cherry-pick results aggregate the best results for each task or instance across historical evaluations.

\begin{table}[H]
\centering
\footnotesize
\setlength{\tabcolsep}{3pt}
\setlength{\abovecaptionskip}{0pt}
\setlength{\belowcaptionskip}{6pt}
\caption{Overall, task-subset, and cherry-pick results. Official 2026 Test results and method names are taken from the \href{https://huggingface.co/spaces/behavior-1k/2026-challenge-leaderboard}{official leaderboard}. The 2025 and 2026 results follow their respective evaluation protocols (Table~\ref{tab:benchmark_scale}).}
\label{tab:routed}
\begin{tabularx}{\textwidth}{@{}>{\hsize=1.15\hsize\linewidth=\hsize\raggedright\arraybackslash}X>{\hsize=.85\hsize\linewidth=\hsize\raggedright\arraybackslash}Xrcrrrr@{}}
\toprule
Team / configuration & Method & Task IDs & Timeout & Val & Val & Test & Test \\
 & & & & Q-score & SR (\%) & Q-score & SR (\%) \\
\midrule
RLC (2025, 1st)~\cite{larchenko2025} & VLA & 0--49 & $2.0\times\bar{T}_{\mathrm{task}}$ & 0.261 & 11.2 & 0.260 & 12.4 \\
Openpi Comet (2025, 2nd)~\cite{bai2025} & VLA & 0--49 & $2.0\times\bar{T}_{\mathrm{task}}$ & 0.183\textsuperscript{a} & 14.4 & 0.251 & 11.4 \\
Galaxea G0.5\textsuperscript{b}~\cite{galaxea2026} & VLA & 0--49 & -- & 0.314 & -- & -- & -- \\
\midrule
The North Star (2026, 1st) & Hiearchical Phys. RSI & 0--99 & $1.5\times\bar{T}_{\mathrm{task}}$ & -- & -- & 0.683 & 48.2 \\
\textbf{IF1001 (Ours, 2nd)} & WAM & 0--99 & $1.5\times\bar{T}_{\mathrm{task}}$ & $0.613 \pm 0.008$\textsuperscript{c} & $40.9 \pm 1.0$ & \textbf{0.615} & \textbf{41.2} \\
Cyber (2026, 3rd) & WAM & 0--99 & $1.5\times\bar{T}_{\mathrm{task}}$ & -- & -- & 0.596 & 39.6 \\
Anyverse (2026, 4th) & A$^2$MEAS & 0--99 & $1.5\times\bar{T}_{\mathrm{task}}$ & -- & -- & 0.498 & 32.0 \\
Autolab (2026, 5th) & A$^2$WLA & 0--99 & $1.5\times\bar{T}_{\mathrm{task}}$ & -- & -- & 0.495 & 30.9 \\
XYRobot (2026, 6th) & VLA+RSI & 0--99 & $1.5\times\bar{T}_{\mathrm{task}}$ & -- & -- & 0.484 & 28.9 \\
\midrule
Ours & WAM & 0--49 & $1.5\times\bar{T}_{\mathrm{task}}$ & $0.588 \pm 0.011$ & $38.1 \pm 1.3$ & -- & -- \\
Ours & WAM & 50--99 & $1.5\times\bar{T}_{\mathrm{task}}$ & $0.639 \pm 0.006$ & $43.8 \pm 0.7$ & -- & -- \\
\midrule
Ours (task cherry-pick) & WAM & 0--99 & $1.5\times\bar{T}_{\mathrm{task}}$ & 0.738 & 55.2 & -- & -- \\
Ours (instance cherry-pick) & WAM & 0--99 & $1.5\times\bar{T}_{\mathrm{task}}$ & 0.891 & 79.5 & -- & -- \\
\bottomrule
\end{tabularx}
\par\vspace{3pt}
\begin{minipage}{\textwidth}
\footnotesize
\textsuperscript{a}Comet's validation evaluation was incomplete at submission; we report the \href{https://behavior.stanford.edu/challenge/archive/2025/leaderboard.html}{official leaderboard} results.
\par
\textsuperscript{b}G0.5 results use a single checkpoint trained for 4 epochs and average 2 runs of 10 instances per task. The timeout multiplier, full-task success rates, and private-test scores are unreported~\cite{galaxea2026}.
\par
\textsuperscript{c}Our Val results evaluate the method presented in this report, which simplifies the deployment configuration used for our official private-test submission. This simplification results in a modest decrease in performance. Our overall and task-subset Val results report mean $\pm$ sample standard deviation across the 3 evaluation rounds in Appendix~\ref{app:per_task}, with tasks weighted equally within each round.
\end{minipage}
\end{table}

Our method, IF1001, ranks second on the official 2026 leaderboard, achieving a test Q-score of 0.615 and a success rate of 41.2\%, without relying on any closed-source tools. During official private-test evaluation, our system served concurrent requests across 100 evaluation ports with limited hardware resources, achieving an average end-to-end inference rate above 2~FPS per port. With a single request at a time, the end-to-end inference rate exceeds 4 FPS, with approximately 56 GB of GPU memory usage.

Under instance-level cherry-picking, our method achieves a Q-score of 0.891 and a success rate of 79.5\%. Although these results represent an optimistic upper bound rather than the performance of a single fixed configuration, they highlight the substantial potential of WAM-based methods and suggest considerable room for further improvement.

\subsection{Ablation study}

\subsubsection{Task-group LoRA fine-tuning}
\label{sec:base}

The base model is trained on demonstrations from all 100 tasks. Starting from this model, we freeze the base weights and train a separate LoRA adapter for each of the 17 task groups for 40,000 steps. Table~\ref{tab:base} shows that this task-group specialization improves performance over the base policy, with the execution window and simulator version held fixed. Appendix~\ref{app:group_policy_comparison} reports the comparison by task group.

\begin{table}[H]
\centering
\small
\setlength{\tabcolsep}{4pt}
\setlength{\abovecaptionskip}{0pt}
\setlength{\belowcaptionskip}{6pt}
\caption{Base policy model with and without task-group LoRA adapters.}
\label{tab:base}
\begin{tabularx}{0.95\textwidth}{@{}Xrrr@{}}
\toprule
Configuration & No. of tasks & Val & Val \\
 & & Q-score & SR (\%) \\
\midrule
Base policy & 100 & 0.434 & 21.3 \\
Base policy + task-group LoRA adapters & 100 & 0.494 & 26.9 \\
\bottomrule
\end{tabularx}
\end{table}

\subsubsection{Effect of stage-ID estimation}
\label{sec:stage_comparison}

Table~\ref{tab:stage_comparison} compares uniform clock-derived stage IDs with stage estimation followed by voting-based post-processing across all 100 tasks. The performance gain highlights the importance of providing the policy with stage IDs that more accurately reflect task progress. Appendix~\ref{app:stage_rules} describes the two rules.

\begin{table}[H]
\centering
\small
\setlength{\tabcolsep}{4pt}
\setlength{\abovecaptionskip}{0pt}
\setlength{\belowcaptionskip}{6pt}
\caption{Predicted versus clock-derived stage IDs for task-group LoRA adapters.}
\label{tab:stage_comparison}
\begin{tabularx}{0.95\textwidth}{@{}Xrrr@{}}
\toprule
Stage-ID strategy & No. of tasks & Val & Val \\
 & & Q-score & SR (\%) \\
\midrule
Uniform episode clock & 100 & 0.480 & 25.2 \\
Stage estimation + post-processing & 100 & 0.494 & 26.9 \\
\bottomrule
\end{tabularx}
\end{table}

\paragraph{Skill-to-stage lookup.}
We evaluate skill-to-stage lookup on the 24 tasks ranked highest by the subtask--stage association score $A_q$ (Appendix~\ref{app:skill_stage_association}). Lookup outperforms stage estimation with post-processing on these tasks (Table~\ref{tab:planner_comparison}). For these tasks, subtask prediction followed by lookup provides stage IDs without a separate stage-estimation branch.

These results motivate exploring joint subtask and stage-ID prediction, or conditioning directly on subtasks to address non-Markovian ambiguity. Human-annotated subtasks provide explicit operation semantics beyond temporal stage labels, but competition time constraints prevented us from evaluating these alternatives.

\begin{table}[H]
\centering
\small
\setlength{\tabcolsep}{4pt}
\setlength{\abovecaptionskip}{0pt}
\setlength{\belowcaptionskip}{6pt}
\caption{Skill-to-stage lookup on 24 tasks with complete results in both configurations.}
\label{tab:planner_comparison}
\begin{tabularx}{0.95\textwidth}{@{}Xrrr@{}}
\toprule
Configuration & No. of tasks & Val & Val \\
 & & Q-score & SR (\%) \\
\midrule
Stage estimation + post-processing & 24 & 0.681 & 52.5 \\
Skill-to-stage lookup & 24 & 0.707 & 56.3 \\
\bottomrule
\end{tabularx}
\end{table}

\subsubsection{Skill-specific LoRA adapters}
\label{sec:skill_comparisons}

After selecting the best task-specific routes from the preceding evaluations, we reevaluate the system with Pick/Place/Close/Open LoRA adapter routing, achieving a Q-score of 0.581 and a success rate of 38.1\% across all 100 tasks. Each skill adapter is trained on demonstration segments pooled across tasks. At inference, the router uses navigation/operation predictions from the progress estimator to determine when skill routing is enabled, and skill predictions to select the appropriate adapter in place of the task-group adapter.

\subsubsection{Temporal retiming and execution windows}
\label{sec:retiming_comparison}

We compare 26-to-20 execution, 52-to-40 execution, and TOPP-RA retiming, following the definitions and bypass and fallback rules in Section~\ref{sec:execution}.

\begin{table}[H]
\centering
\small
\setlength{\tabcolsep}{4pt}
\setlength{\abovecaptionskip}{0pt}
\setlength{\belowcaptionskip}{6pt}
\caption{Task performance and policy-call cost of execution-window and retiming strategies.}
\label{tab:window_comparison}
\begin{tabularx}{0.95\textwidth}{@{}Xrrrr@{}}
\toprule
Configuration & Model & No. of tasks & Val & Val \\
 & calls (k) & & Q-score & SR (\%) \\
\midrule
Short-window & 676.7 & 100 & 0.575 & 35.9 \\
Long-window & 340.0 & 100 & 0.549 & 33.5 \\
TOPP-RA & 540.3 & 100 & 0.557 & 36.3 \\
\bottomrule
\end{tabularx}
\par\smallskip
\begin{minipage}{0.95\textwidth}
\footnotesize
Model calls denote the number of times Cosmos is invoked to generate an action chunk.
\end{minipage}
\end{table}

Table~\ref{tab:window_comparison} shows a performance--cost trade-off. Short-window execution achieves the highest Q-score but requires the most model calls. Switching from 26-to-20 to 52-to-40 execution reduces calls by 49.8\% while reducing task performance. TOPP-RA uses 20.2\% fewer calls than short-window execution and improves performance over long-window execution.

Table~\ref{tab:toppra_time} compares simulated execution time on 260 instances across 60 tasks that succeeded under both TOPP-RA and uniform temporal resampling. The baseline uses uniform temporal resampling with task-specific 26-to-20 or 52-to-40 execution windows. We first average these instances within each task, then average the 60 task means equally. The human reference averages the official mean demonstration durations for the same tasks. This measures task execution speed, not single-call model latency.

\begin{table}[H]
\centering
\small
\setlength{\tabcolsep}{4pt}
\setlength{\abovecaptionskip}{0pt}
\setlength{\belowcaptionskip}{6pt}
\caption{Task-averaged execution time across 60 tasks. Robot results use 260 instances successful under both configurations; the human reference uses the same tasks.}
\label{tab:toppra_time}
\begin{tabularx}{0.95\textwidth}{@{}Xr@{}}
\toprule
Configuration & Mean time (s) \\
\midrule
Human demonstrations & 313.12 \\
Uniform temporal resampling & 274.57 \\
TOPP-RA & 211.40 \\
\bottomrule
\end{tabularx}
\end{table}

\section{Discussion and Future Directions}
{ We first thank the organizers of the 2026 BEHAVIOR Challenge for providing an excellent platform. We learned a great deal from participating in this competition.

Our main reflections after this competition are as follows:

\begin{enumerate}
    \item \textbf{Recursive self-improvement offers a promising direction.}
    RSI has led us to think about a possible ``GPT moment'' for embodied intelligence. We view large language models as the brain of embodied agents and world action models (WAMs) as the cerebellum. Improving either component can improve the overall system, but how to coordinate them effectively remains an open question.

    \item \textbf{WAMs are capable.}
    In our early experiments, we explored VLA approaches based on $\pi_{0.5}$. We observed a substantial performance improvement after switching to WAMs, which convinced us of their capability.

    \item \textbf{WAMs have considerable potential.}
    Our official test Q-score is 0.615. However, when we run the same task or instance multiple times under different configurations and select the best result, instance-level cherry-picking yields a Q-score close to 0.9. This gap suggests that the current method is capable of producing strong results, although it cannot yet do so consistently.

    \item \textbf{Scaling up is the near future.}
    The Bitter Lesson~\cite{sutton2019} suggests that simple and scalable methods with fewer human priors are more likely to benefit from increased computation. Our computational resources were limited during this competition, so we could not fully explore scaling. We expect further gains from larger models, more data, and longer training. For example, Cosmos 3 includes an understanding model and a generation model. Since the instruction-following and spatial-understanding abilities of the generation model largely come from the understanding model, we expect that scaling the understanding model will further improve the generation model.

\end{enumerate}

}

\subsection*{AI use statement}
{ We used generative AI tools to assist with manuscript writing and editing. We provided the AI with the structure of the content we wanted to convey and a draft, let the AI help us rewrite and polish it, and then edited it ourselves.}

\newpage 

\appendix
\makeatletter
\def\appendixprefix@section{Appendix\ }
\renewcommand{\@seccntformat}[1]{\csname appendixprefix@#1\endcsname\csname the#1\endcsname\quad}
\makeatother

\section{Task-group initial observations}
\label{app:views}

Each panel shows a training episode's initial head-camera view above synchronized wrist views. Groups 1--15 contain 6 tasks each; groups 16--17 contain 5 each.

\begin{figure}[H]
\centering
\reportfig[width=\textwidth,height=0.82\textheight,keepaspectratio]{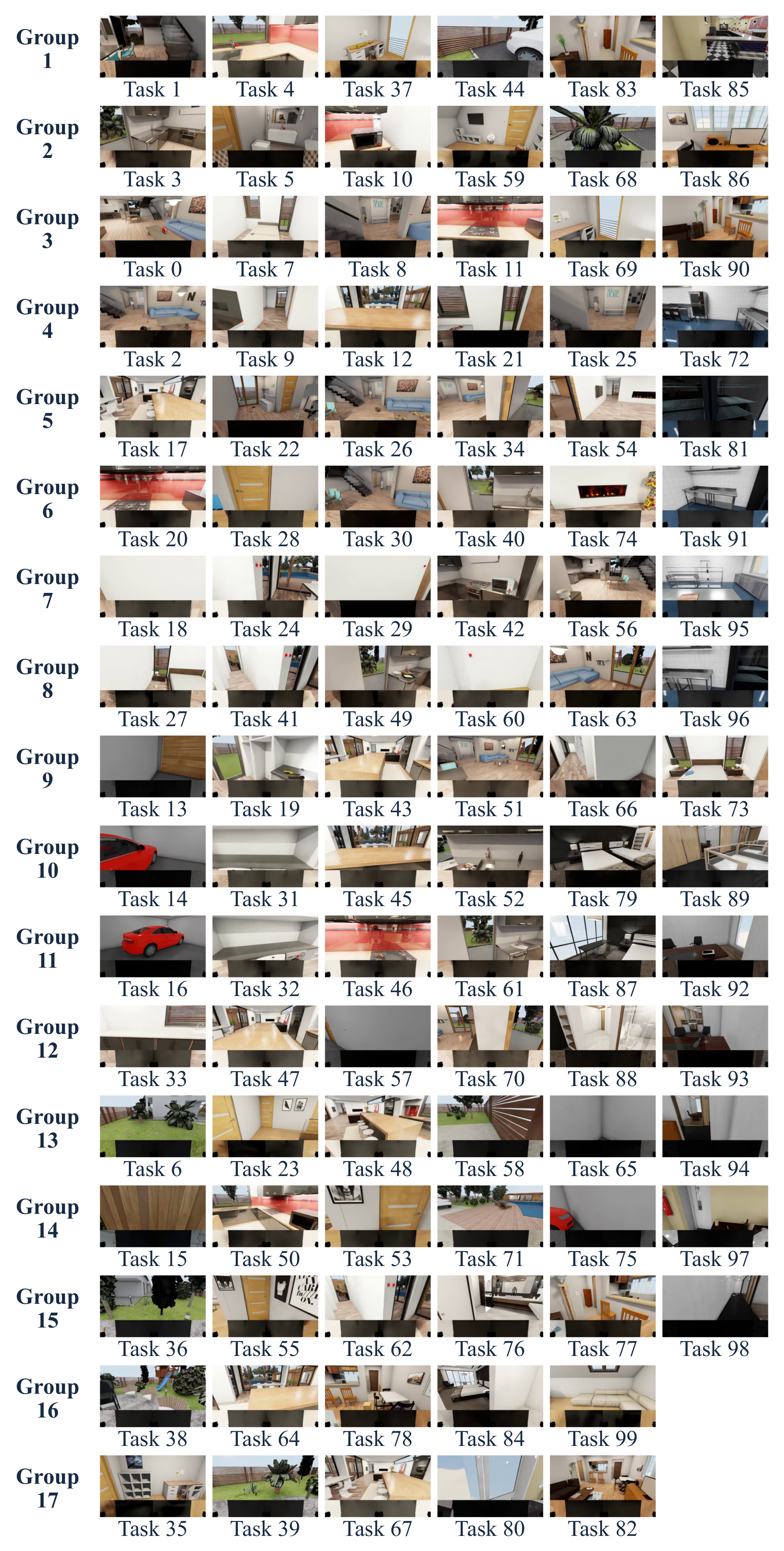}
\caption{Initial observations for the 17 task groups.}
\label{fig:groups}
\end{figure}

\clearpage
\section{Visual examples of skill-specific LoRA adapters}
\label{app:skillviews}

\begin{figure}[H]
\centering
\reportfig[width=0.92\textwidth,height=0.72\textheight,keepaspectratio]{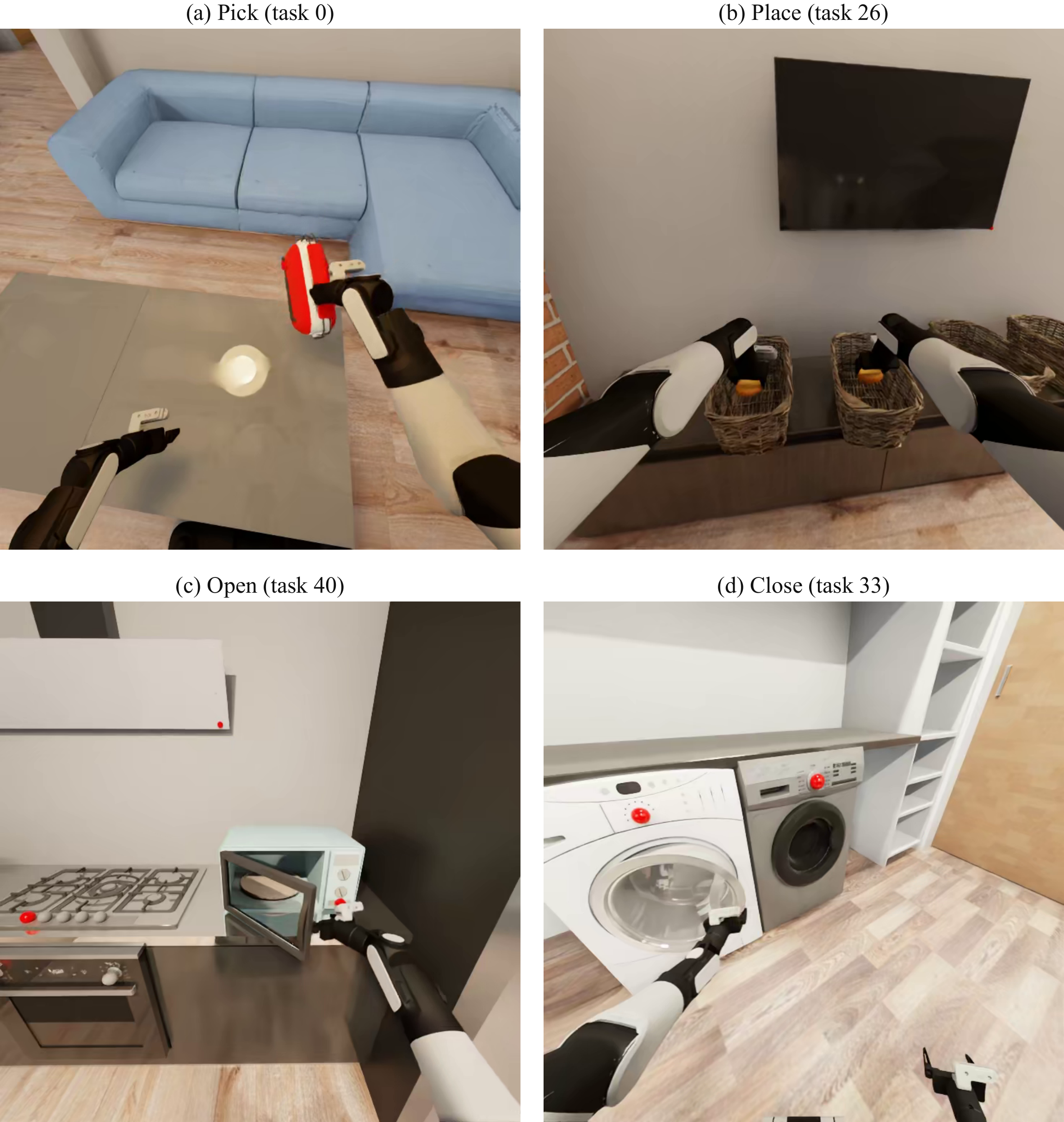}
\caption{Head-camera views during demonstrated skill execution: (a) picking up a radio, (b) placing a cookie in a basket, (c) opening a microwave door, and (d) closing a washing-machine door.}
\label{fig:skillviews}
\end{figure}

Figure~\ref{fig:skillviews} illustrates the 4 skill families with training frames from 4 different tasks; Table~\ref{tab:skills} lists the corresponding skill categories.

\clearpage
\section{Detailed validation results}
\label{app:evaluation}

Rollout videos are available on Hugging Face:
\begin{itemize}
\item Self-evaluation (Val): \url{https://huggingface.co/datasets/YelinHe/behavior_2026_self_evaluation_simplify}
\item Official private-test evaluation (Test): \nolinkurl{https://huggingface.co/datasets/YelinHe/behavior_2026_official_private_test}
\end{itemize}

\subsection{Per-task validation results}
\label{app:per_task}

Table~\ref{tab:per_task} reports per-task validation results for 3 evaluation rounds, each using 10 instances per task.

\begingroup
\footnotesize
\setlength{\tabcolsep}{3.5pt}
\setlength{\LTpre}{10pt}
\setlength{\LTpost}{0pt}
\setcounter{LTchunksize}{200}
\renewcommand{\arraystretch}{1.02}
\begin{longtable}{@{}r l *{6}{r}@{}}
\caption{Per-task validation results across 3 evaluation rounds (100 tasks, 10 instances per task per round).}\label{tab:per_task}\\
\toprule
Task ID & Task & \multicolumn{2}{c}{Val: Round 0} & \multicolumn{2}{c}{Val: Round 1} & \multicolumn{2}{c}{Val: Round 2} \\
\cmidrule(lr){3-4}\cmidrule(lr){5-6}\cmidrule(l){7-8}
 & & Q-score & SR (\%) & Q-score & SR (\%) & Q-score & SR (\%) \\
\midrule
\endfirsthead
\multicolumn{8}{c}{\tablename~\thetable{} (continued)}\\
\toprule
Task ID & Task & \multicolumn{2}{c}{Val: Round 0} & \multicolumn{2}{c}{Val: Round 1} & \multicolumn{2}{c}{Val: Round 2} \\
\cmidrule(lr){3-4}\cmidrule(lr){5-6}\cmidrule(l){7-8}
 & & Q-score & SR (\%) & Q-score & SR (\%) & Q-score & SR (\%) \\
\midrule
\endhead
\midrule
\multicolumn{8}{r}{Continued on next page}\\
\endfoot
\bottomrule
\endlastfoot
0 & Turning on radio & 0.700 & 70.0 & 0.700 & 70.0 & 0.600 & 60.0 \\
1 & Picking up trash & 1.000 & 100.0 & 1.000 & 100.0 & 1.000 & 100.0 \\
2 & Putting away Halloween decorations & 0.800 & 40.0 & 0.886 & 60.0 & 0.800 & 40.0 \\
3 & Cleaning up plates and food & 0.586 & 40.0 & 0.643 & 50.0 & 0.500 & 30.0 \\
4 & Can meat & 0.211 & 0.0 & 0.356 & 20.0 & 0.322 & 10.0 \\
5 & Setting mousetraps & 0.967 & 90.0 & 1.000 & 70.0 & 0.983 & 90.0 \\
6 & Hiding Easter eggs & 0.278 & 20.0 & 0.300 & 20.0 & 0.267 & 10.0 \\
7 & Picking up toys & 0.500 & 0.0 & 0.433 & 0.0 & 0.417 & 0.0 \\
8 & Rearranging kitchen furniture & 0.725 & 40.0 & 0.750 & 30.0 & 0.750 & 30.0 \\
9 & Putting up Christmas decorations inside & 0.533 & 0.0 & 0.500 & 0.0 & 0.422 & 0.0 \\
10 & Set up a coffee station in your kitchen & 0.383 & 0.0 & 0.383 & 0.0 & 0.433 & 10.0 \\
11 & Putting dishes away after cleaning & 0.914 & 80.0 & 1.000 & 100.0 & 0.907 & 80.0 \\
12 & Preparing lunch box & 0.700 & 20.0 & 0.650 & 30.0 & 0.800 & 50.0 \\
13 & Loading the car & 0.833 & 70.0 & 0.867 & 60.0 & 0.967 & 90.0 \\
14 & Carrying in groceries & 0.725 & 60.0 & 0.825 & 70.0 & 0.725 & 60.0 \\
15 & Bringing in wood & 0.967 & 90.0 & 0.967 & 90.0 & 0.933 & 80.0 \\
16 & Moving boxes to storage & 1.000 & 100.0 & 0.900 & 90.0 & 0.900 & 90.0 \\
17 & Bringing water & 0.500 & 50.0 & 0.933 & 90.0 & 0.900 & 90.0 \\
18 & Tidying bedroom & 0.833 & 60.0 & 0.667 & 50.0 & 0.833 & 60.0 \\
19 & Outfit a basic toolbox & 0.400 & 10.0 & 0.386 & 20.0 & 0.514 & 40.0 \\
20 & Sorting vegetables & 0.731 & 0.0 & 0.669 & 10.0 & 0.731 & 0.0 \\
21 & Collecting children's toys & 0.814 & 30.0 & 0.843 & 30.0 & 0.871 & 50.0 \\
22 & Putting shoes on rack & 0.920 & 60.0 & 0.950 & 80.0 & 0.880 & 60.0 \\
23 & Boxing books up for storage & 0.583 & 10.0 & 0.750 & 30.0 & 0.617 & 20.0 \\
24 & Storing food & 0.825 & 40.0 & 0.887 & 40.0 & 0.787 & 30.0 \\
25 & Clearing food from table into fridge & 0.660 & 40.0 & 0.700 & 50.0 & 0.760 & 50.0 \\
26 & Assembling gift baskets & 0.750 & 20.0 & 0.750 & 0.0 & 0.625 & 0.0 \\
27 & Sorting household items & 0.637 & 20.0 & 0.487 & 10.0 & 0.562 & 20.0 \\
28 & Getting organized for work & 0.100 & 0.0 & 0.070 & 0.0 & 0.070 & 0.0 \\
29 & Clean up your desk & 0.355 & 0.0 & 0.318 & 0.0 & 0.409 & 0.0 \\
30 & Setting the fire & 0.637 & 10.0 & 0.588 & 10.0 & 0.675 & 20.0 \\
31 & Clean boxing gloves & 0.950 & 90.0 & 1.000 & 100.0 & 0.950 & 90.0 \\
32 & Wash a baseball cap & 0.800 & 80.0 & 0.600 & 60.0 & 0.800 & 80.0 \\
33 & Wash dog toys & 0.200 & 20.0 & 0.400 & 40.0 & 0.600 & 60.0 \\
34 & Hanging pictures & 0.500 & 50.0 & 0.600 & 60.0 & 0.300 & 30.0 \\
35 & Attach a camera to a tripod & 0.100 & 10.0 & 0.100 & 10.0 & 0.000 & 0.0 \\
36 & Clean a patio & 0.100 & 10.0 & 0.000 & 0.0 & 0.000 & 0.0 \\
37 & Clean a trumpet & 0.000 & 0.0 & 0.000 & 0.0 & 0.000 & 0.0 \\
38 & Spraying for bugs & 0.350 & 20.0 & 0.400 & 20.0 & 0.350 & 20.0 \\
39 & Spraying fruit trees & 0.400 & 30.0 & 0.600 & 50.0 & 0.600 & 60.0 \\
40 & Make microwave popcorn & 1.000 & 100.0 & 1.000 & 100.0 & 0.900 & 90.0 \\
41 & Cook cabbage & 0.600 & 40.0 & 0.350 & 10.0 & 0.600 & 40.0 \\
42 & Chop an onion & 0.450 & 30.0 & 0.725 & 50.0 & 0.825 & 60.0 \\
43 & Slicing vegetables & 0.733 & 50.0 & 0.511 & 30.0 & 0.622 & 40.0 \\
44 & Chopping wood & 0.350 & 10.0 & 0.350 & 0.0 & 0.350 & 0.0 \\
45 & Cook hot dogs & 0.900 & 90.0 & 0.900 & 80.0 & 0.900 & 90.0 \\
46 & Cook bacon & 0.229 & 10.0 & 0.671 & 30.0 & 0.443 & 10.0 \\
47 & Freeze pies & 0.471 & 20.0 & 0.257 & 10.0 & 0.271 & 0.0 \\
48 & Canning food & 0.100 & 0.0 & 0.240 & 10.0 & 0.130 & 0.0 \\
49 & Make pizza & 0.000 & 0.0 & 0.000 & 0.0 & 0.000 & 0.0 \\
50 & Freeze fruit & 0.414 & 20.0 & 0.429 & 10.0 & 0.486 & 0.0 \\
51 & Cook a brisket & 0.533 & 40.0 & 0.733 & 70.0 & 0.800 & 70.0 \\
52 & Sorting bottles cans and paper & 0.319 & 10.0 & 0.225 & 0.0 & 0.306 & 10.0 \\
53 & Tidying living room & 0.625 & 30.0 & 0.700 & 30.0 & 0.525 & 0.0 \\
54 & Putting away toys & 0.900 & 40.0 & 0.887 & 40.0 & 0.887 & 50.0 \\
55 & Reshelving library books & 0.933 & 80.0 & 0.867 & 70.0 & 0.933 & 80.0 \\
56 & Make rose centerpieces & 0.875 & 70.0 & 0.950 & 80.0 & 0.875 & 50.0 \\
57 & Sweeping garage & 0.750 & 60.0 & 0.650 & 50.0 & 0.650 & 40.0 \\
58 & Stacking wood & 0.650 & 30.0 & 0.800 & 40.0 & 0.700 & 50.0 \\
59 & Organizing art supplies & 0.640 & 30.0 & 0.640 & 20.0 & 0.740 & 40.0 \\
60 & Scrubbing bathroom floor & 0.800 & 80.0 & 0.500 & 50.0 & 0.600 & 60.0 \\
61 & Bringing paper to recycling & 0.267 & 20.0 & 0.333 & 20.0 & 0.267 & 20.0 \\
62 & Halve an egg & 0.520 & 30.0 & 0.700 & 50.0 & 0.500 & 20.0 \\
63 & Installing smoke detectors & 0.000 & 0.0 & 0.000 & 0.0 & 0.200 & 20.0 \\
64 & Setting the table & 0.650 & 20.0 & 0.700 & 20.0 & 0.662 & 30.0 \\
65 & Unloading the car & 0.900 & 90.0 & 0.700 & 60.0 & 1.000 & 100.0 \\
66 & Turning out all lights before sleep & 0.640 & 20.0 & 0.800 & 40.0 & 0.840 & 50.0 \\
67 & Boxing food after dinner & 0.600 & 40.0 & 0.600 & 50.0 & 0.617 & 40.0 \\
68 & Cleaning up branches and twigs & 0.320 & 0.0 & 0.260 & 0.0 & 0.300 & 0.0 \\
69 & Vacuuming floors & 0.200 & 20.0 & 0.100 & 10.0 & 0.000 & 0.0 \\
70 & Thawing frozen food & 0.556 & 30.0 & 0.522 & 20.0 & 0.533 & 30.0 \\
71 & Clean your rusty garden tools & 0.240 & 0.0 & 0.340 & 10.0 & 0.240 & 0.0 \\
72 & Cook a frozen pie & 0.750 & 70.0 & 0.800 & 80.0 & 0.700 & 70.0 \\
73 & Organizing school stuff & 0.817 & 20.0 & 0.883 & 70.0 & 0.850 & 30.0 \\
74 & Carrying out garden furniture & 0.900 & 90.0 & 0.750 & 70.0 & 0.950 & 90.0 \\
75 & Put together a basic pruning kit & 0.825 & 80.0 & 0.825 & 80.0 & 0.725 & 60.0 \\
76 & Dispose of glass & 0.875 & 60.0 & 0.975 & 90.0 & 1.000 & 100.0 \\
77 & Installing a modem & 0.700 & 50.0 & 0.700 & 40.0 & 0.750 & 60.0 \\
78 & Make cabinet doors & 0.200 & 20.0 & 0.100 & 10.0 & 0.100 & 10.0 \\
79 & Polishing shoes & 0.367 & 0.0 & 0.333 & 0.0 & 0.333 & 0.0 \\
80 & Clean up broken glass & 0.767 & 50.0 & 0.800 & 60.0 & 0.633 & 30.0 \\
81 & Packing meal for delivery & 0.617 & 10.0 & 0.683 & 40.0 & 0.667 & 20.0 \\
82 & Store batteries & 0.867 & 80.0 & 0.867 & 80.0 & 0.867 & 80.0 \\
83 & Store honey & 1.000 & 100.0 & 1.000 & 100.0 & 1.000 & 100.0 \\
84 & Tidying bathroom & 0.875 & 60.0 & 0.900 & 70.0 & 0.825 & 60.0 \\
85 & Putting dirty dishes in sink & 0.775 & 50.0 & 0.775 & 50.0 & 0.825 & 60.0 \\
86 & Make gift bags for baby showers & 0.433 & 10.0 & 0.500 & 0.0 & 0.600 & 20.0 \\
87 & Collecting aluminum cans & 0.883 & 60.0 & 0.933 & 70.0 & 0.750 & 60.0 \\
88 & Rearrange your room & 0.767 & 70.0 & 0.933 & 90.0 & 0.967 & 90.0 \\
89 & Installing a fax machine & 0.550 & 50.0 & 0.500 & 50.0 & 0.700 & 60.0 \\
90 & Composting waste & 0.900 & 80.0 & 0.950 & 90.0 & 0.950 & 90.0 \\
91 & Store produce & 0.900 & 80.0 & 0.825 & 60.0 & 0.825 & 60.0 \\
92 & Installing a scanner & 0.350 & 30.0 & 0.150 & 10.0 & 0.350 & 20.0 \\
93 & Clean a keyboard & 0.400 & 40.0 & 0.400 & 40.0 & 0.400 & 40.0 \\
94 & Dispose of batteries & 0.900 & 80.0 & 0.650 & 30.0 & 0.750 & 50.0 \\
95 & Cook Brussels sprouts & 0.323 & 0.0 & 0.446 & 10.0 & 0.269 & 0.0 \\
96 & Cook broccolini & 0.818 & 40.0 & 0.864 & 70.0 & 0.745 & 40.0 \\
97 & Set up a bar for a cocktail party & 0.825 & 30.0 & 0.840 & 30.0 & 0.795 & 30.0 \\
98 & Laying tile floors & 0.863 & 80.0 & 0.875 & 80.0 & 0.975 & 90.0 \\
99 & Sorting books on shelf & 0.091 & 0.0 & 0.164 & 10.0 & 0.291 & 20.0 \\
\midrule
\multicolumn{2}{@{}l}{\textbf{Overall}} & \textbf{0.605} & \textbf{39.8} & \textbf{0.617} & \textbf{41.6} & \textbf{0.619} & \textbf{41.4} \\
\end{longtable}
\endgroup

\clearpage

\subsection{Task-group LoRA fine-tuning by group}
\label{app:group_policy_comparison}

Table~\ref{tab:group_policy_comparison} presents the results from Table~\ref{tab:base} for each of the 17 task groups defined in Table~\ref{tab:groups}. Task-group LoRA fine-tuning improves Q-score in 14 groups but lowers it in groups 3, 12, and 16.

\begin{table}[H]
\centering
\small
\setlength{\tabcolsep}{4pt}
\setlength{\abovecaptionskip}{0pt}
\setlength{\belowcaptionskip}{6pt}
\caption{Base policy versus task-group LoRA fine-tuning within each task group.}
\label{tab:group_policy_comparison}
\begin{tabular*}{0.95\textwidth}{@{\extracolsep{\fill}}rrrrrr@{}}
\toprule
Group & No. of tasks & \multicolumn{2}{c}{Base policy} & \multicolumn{2}{c}{Task-group LoRA} \\
\cmidrule(lr){3-4}\cmidrule(l){5-6}
 & & Val & Val & Val & Val \\
 & & Q-score & SR (\%) & Q-score & SR (\%) \\
\midrule
1 & 6 & 0.485 & 38.3 & 0.509 & 38.3 \\
2 & 6 & 0.507 & 20.0 & 0.559 & 21.7 \\
3 & 6 & 0.380 & 20.0 & 0.351 & 18.3 \\
4 & 6 & 0.575 & 15.0 & 0.626 & 30.0 \\
5 & 6 & 0.495 & 16.7 & 0.654 & 31.7 \\
6 & 6 & 0.554 & 38.3 & 0.577 & 38.3 \\
7 & 6 & 0.435 & 6.7 & 0.548 & 16.7 \\
8 & 6 & 0.291 & 16.7 & 0.305 & 20.0 \\
9 & 6 & 0.528 & 30.0 & 0.635 & 35.0 \\
10 & 6 & 0.461 & 30.0 & 0.568 & 40.0 \\
11 & 6 & 0.367 & 20.0 & 0.380 & 20.0 \\
12 & 6 & 0.362 & 21.7 & 0.276 & 18.3 \\
13 & 6 & 0.338 & 13.3 & 0.419 & 18.3 \\
14 & 6 & 0.348 & 8.3 & 0.583 & 23.3 \\
15 & 6 & 0.597 & 36.7 & 0.628 & 43.3 \\
16 & 5 & 0.261 & 6.0 & 0.200 & 4.0 \\
17 & 5 & 0.350 & 22.0 & 0.540 & 38.0 \\
\midrule
Overall & 100 & 0.434 & 21.3 & 0.494 & 26.9 \\
\bottomrule
\end{tabular*}
\end{table}

\clearpage
\section{Stage-ID post-processing rules}
\label{app:stage_rules}

For task $q$, $K_q$ is the stage count defined in Eq.~\eqref{eq:stage_labels}. Let $\hat s$ denote the raw predicted stage ID and $\tilde s$ the stage ID used for policy conditioning, replacing the training label $s$ in Eq.~\eqref{eq:stage_tokens}. The buffer $\mathcal H$ stores up to 3 recent predictions clipped to $[0,K_q-1]$, and $C(j)$ counts occurrences of stage $j$ in the buffer, with $C(j)=0$ outside this range. Initialize $\tilde s=0$ and $\mathcal H=\varnothing$ at episode reset. The voting update runs once per new action chunk; $\tilde s_{\mathrm{old}}$ stores the stage ID before the update.

\par\medskip
\noindent\begin{minipage}{\linewidth}
\small
\textbf{Stage vote} $\operatorname{Vote}(\tilde s,\mathcal H,\hat s)$:
\begin{tabbing}
\quad\=\quad\=\kill
\>Append $\operatorname{clip}(\hat s,0,K_q-1)$ to $\mathcal H$; retain the latest 3.\\
\>\textbf{if} $|\mathcal H|<3$: \textbf{return} $\tilde s$\\
\>$\tilde s_{\mathrm{old}}\leftarrow \tilde s$\\
\>\textbf{if} $\tilde s<K_q-1$ \textbf{and} $(C(\tilde s+1)\geq2\ \textbf{or}\ C(\tilde s+2)=3)$:\\
\>\>$\tilde s\leftarrow \tilde s+1$\\
\>\textbf{else if} $\tilde s>0$ \textbf{and} $C(\tilde s-1)=3$ \textbf{and} $\tilde s<K_q-1$:\\
\>\>$\tilde s\leftarrow \tilde s-1$\\
\>\textbf{if} $\tilde s\neq \tilde s_{\mathrm{old}}$: clear $\mathcal H$\\
\>\textbf{return} $\tilde s$
\end{tabbing}
\end{minipage}

\par\medskip
\noindent\textbf{Uniform episode clock.} Let $n_{\mathrm{sim}}$ be the elapsed simulator steps and $N_{\mathrm{timeout},q}$ the 2026 task timeout in Table~\ref{tab:benchmark_scale} converted to simulator steps. Both use the $f_{\mathrm{sim}}=30$~Hz control grid defined in Section~\ref{sec:execution}. The clock-derived stage is
\begin{equation}
c_q=\operatorname{clip}\!\left(\left\lfloor\frac{K_q n_{\mathrm{sim}}}{N_{\mathrm{timeout},q}}\right\rfloor,0,K_q-1\right).
\label{eq:uniform_episode_clock}
\end{equation}
Set $\tilde s\leftarrow c_q$ without using model predictions.

\clearpage
\section{Subtask--stage association for skill-to-stage lookup}
\label{app:skill_stage_association}

\begin{figure}[H]
\centering
\reportfig[width=0.90\textwidth]{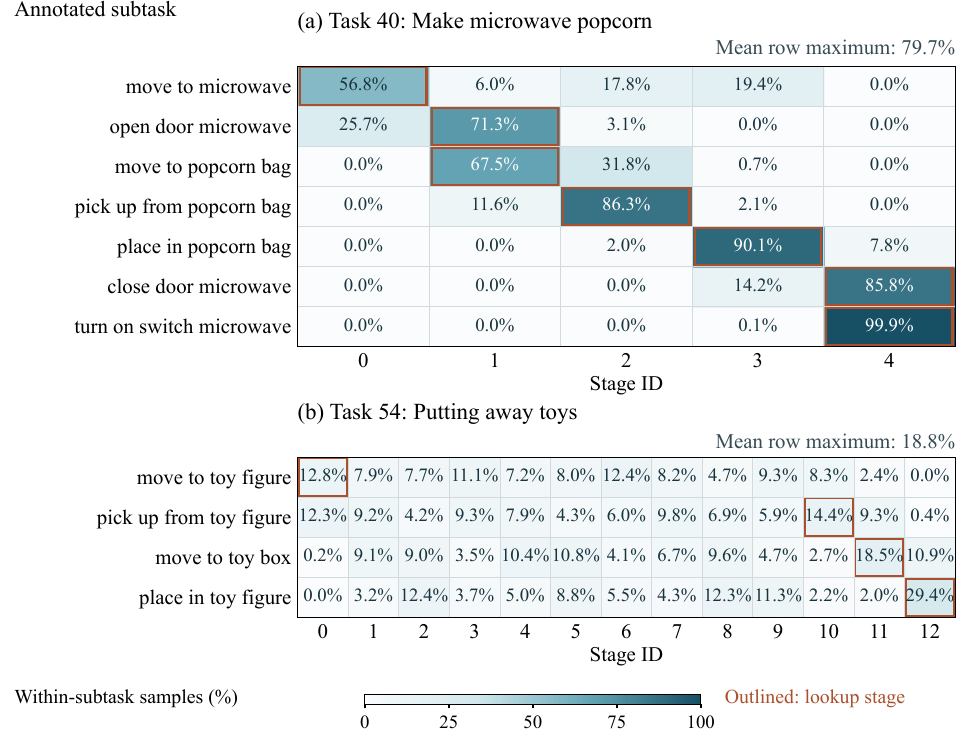}
\caption{Subtask--stage associations in demonstration data: (a) concentrated stage distributions in task 40 and (b) dispersed distributions in task 54. Each row shows the percentage of samples for a given subtask occurring at each stage. Darker cells indicate larger percentages, and outlined cells identify the stage selected by lookup. The mean row maximum summarizes concentration across subtasks. Row totals may differ from 100\% due to rounding.}
\label{fig:skill_stage_association}
\end{figure}

For each task $q$, we sample 200 demonstrations at 1~Hz. Let $C_q(z,s)$ count samples with annotated subtask $z$ and temporal stage $s$ (Eq.~\eqref{eq:stage_labels}), and $\mathcal Z_q$ contain valid, observed subtask labels. We compute
\begin{equation}
P_q(s\mid z)=\frac{C_q(z,s)}{\sum_{s'=0}^{K_q-1}C_q(z,s')},\qquad
A_q=\frac{1}{|\mathcal Z_q|}\sum_{z\in\mathcal Z_q}\max_s P_q(s\mid z).
\label{eq:skill_stage_association}
\end{equation}
Here, $P_q(s\mid z)$ is the stage distribution for subtask $z$, and $A_q$ averages its peak probability across subtasks. Lookup assigns each subtask its most frequent stage. Figure~\ref{fig:skill_stage_association} contrasts task 40's concentrated associations with task 54's repeated subtasks spread across stages.

\end{document}